\documentclass[a4paper,fleqn]{cas-sc}

\usepackage[authoryear]{natbib}
\usepackage{amsmath,amssymb}
\usepackage{booktabs}
\usepackage{graphicx}
\usepackage{subcaption}
\usepackage{stfloats}

\usepackage{float}

\usepackage{multirow}
\usepackage{xcolor}
\usepackage{hyperref}
\usepackage{mathtools}
\usepackage{siunitx}
\usepackage{xspace}
\usepackage{algpseudocode}
\usepackage{longtable}

\usepackage[T1]{fontenc}
\usepackage[utf8]{inputenc}

\providecommand{\bpk}{\Delta\text{Risk-AP}/\text{KB}\xspace}

\begin{document}

\author[1]{Jiaxi Liu}[orcid=0009-0001-2749-6435]
\author[1]{Chengyuan Ma*}[orcid=0000-0002-6337-0450]
\author[1]{Hang Zhou}[orcid=0000-0003-3286-341X]
\author[1]{Weizhe Tang}
\author[1]{Shixiao Liang}
\author[2]{Haoyang Ding}
\author[1]{Xiaopeng Li}[orcid=0000-0002-5264-3775]
\author[1]{Bin Ran}

\affiliation[1]{organization={Department of Civil and Environmental Engineering, University of Wisconsin-Madison},
    city={Madison},
    postcode={53706}, 
    state={Wisconsin},
    country={United States}}
\affiliation[2]{organization={School of Computer, Data and Information Sciences, University of Wisconsin-Madison},
    city={Madison},
    postcode={53706}, 
    state={Wisconsin},
    country={United States}}

\cortext[cma97@wisc.edu]{Corresponding author}

\title [mode = title]{SRA-CP: Spontaneous Risk-Aware Selective Cooperative Perception}

\shorttitle{SRA-CP: Spontaneous Risk-Aware Selective Cooperative Perception}
\shortauthors{Jiaxi Liu}


\maketitle

\begin{abstract}
Cooperative perception (CP) offers significant potential to overcome the limitations of single-vehicle sensing by enabling information sharing among connected vehicles (CVs). However, existing generic CP approaches need to transmit large volumes of perception data that are irrelevant to the driving safety, exceeding available communication bandwidth. Moreover, most CP frameworks rely on pre-defined communication partners, making them unsuitable for dynamic traffic environments. This paper proposes a \textit{Spontaneous Risk-Aware Selective Cooperative Perception (SRA-CP)} framework to address these challenges. SRA-CP introduces a decentralized protocol where connected agents continuously broadcast lightweight perception coverage summaries and initiate targeted cooperation only when risk-relevant blind zones are detected. A perceptual risk identification module enables each CV to locally assess the impact of occlusions on its driving task and determine whether cooperation is necessary. When CP is triggered, the ego vehicle selects appropriate peers based on shared perception coverage and engages in selective information exchange through a fusion module that prioritizes safety-critical content and adapts to bandwidth constraints. We evaluate SRA-CP on a public dataset against several representative baselines. Results show that SRA-CP achieves less than 1\% average precision (AP) loss for safety-critical objects compared to generic CP, while using only 20\% of the communication bandwidth. Moreover, it improves the perception performance by 15\% over existing selective CP methods that do not incorporate risk awareness.

\end{abstract}

\section{Introduction}

Multi-agent cooperative perception (CP) has emerged as a promising paradigm to overcome the limitations of single-agent sensing by enabling agents to share information with each other. In road traffic environments, a single vehicle’s sensing capability is often obstructed by occlusions, resulting in blind zones that lead to hesitation in decision-making and increased collision risk with surrounding participants. These issues are particularly pronounced in scenarios such as unprotected left turns or pedestrians suddenly appearing from behind parked vehicles. With the rapid advancement of connected and automated vehicle (CAV) technologies, ensuring safe and complete perception becomes even more critical, especially for autonomous driving in complex environments \citep{zha2025heterogeneous}. In such contexts, CP offers valuable potential to enhance safety.

\begin{figure}
    \centering
    \includegraphics[width=1\linewidth]{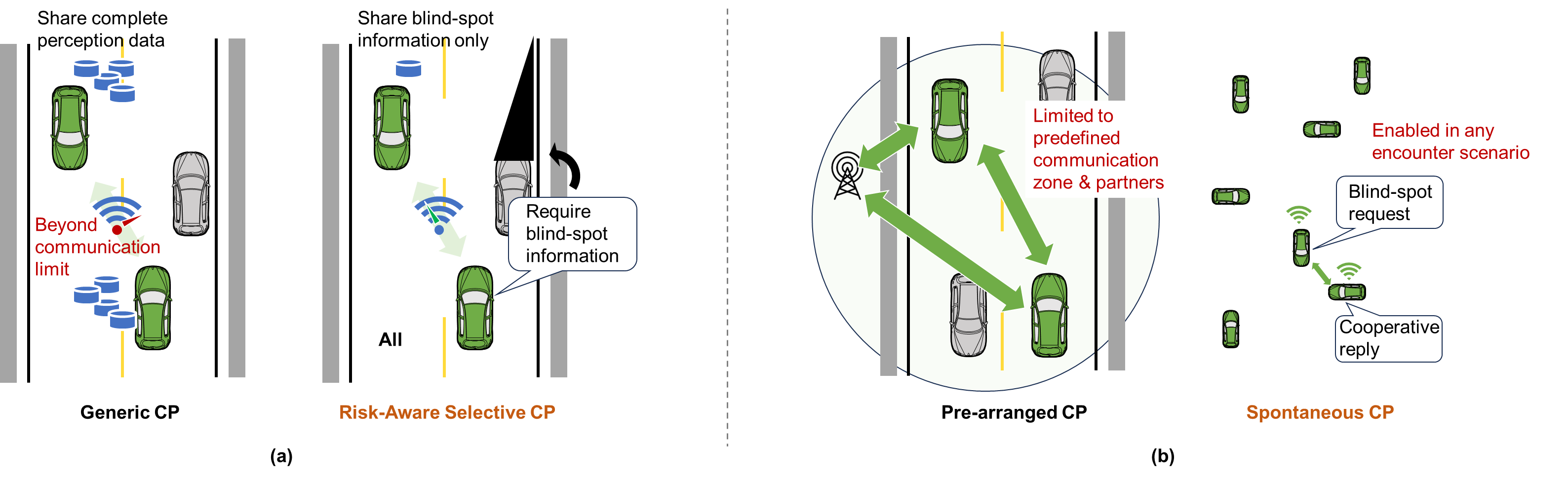}
    \caption{Comparison between \textbf{(a)} \textbf{Generic CP} with full-time information exchange VS the proposed \textbf{Risk-aware selective CP} activated by risky blind-spot events; and \textbf{(b)} \textbf{Pre-arranged CP} constrained by predefined communication partners VS \textbf{Spontaneous CP} enabling dynamic ad-hoc cooperation in arbitrary encounter situations. } 
    \label{fig:intro}
\end{figure}

However, most existing CP studies remain limited to simulations or small-scale experimental setups conducted under ideal and controlled conditions. Achieving large-scale deployment of CP in real-world traffic still faces two major challenges. The first challenge lies in the gap between the massive volume of sensing data generated by CAVs and the limited bandwidth of vehicular communication networks~\citep{hu2022where2comm}. For example, intermediate features extracted from onboard sensors at a rate of 5–20 Hz can produce up to 2MB per frame, translating to a potential transmission rate of 300 Mbps. Such data loads are far beyond what even advanced wireless systems (e.g., 5G) can support in dense environments, especially when vehicles attempt to transmit full perception data simultaneously, as the \textbf{Generic CP} shown in Figure~\ref{fig:intro} (a). Moreover, the problem is exacerbated in real-world traffic, where the number of dynamic agents is large and constantly changing. If every pair of agents were required to maintain real-time communication, the bandwidth burden would grow quadratically with the number of agents. In reality, most of the information being shared is unnecessary—an individual vehicle’s local perception is often sufficient for safe driving in the majority of situations. Even in CP-required cases, not all detected elements need to be transmitted. A more efficient strategy is to share only the information that is both unseen by the receiving vehicle and potentially hazardous to its driving decisions. This observation motivates the concept of \textbf{Risk-aware selective CP}, where each vehicle evaluates the risk level of objects it perceives and only transmits those that satisfy two conditions simultaneously: (i) the object lies within another vehicle’s blind zone, and (ii) the object poses a potential safety risk. For instance, in Figure~\ref{fig:intro}, a parked roadside vehicle obstructs a portion of the scene, and the oncoming vehicle provides supplementary information to complete perception. While recent studies have explored selective CP strategies that only share blind-zone content—resulting in significant communication reduction compared to generic CP \citep{qiu2025map4comm}—they still do not account for the traffic risk relevance of shared content. Our previous work validated that only a small fraction (0.1\%) of driving scenarios actually require CP \citep{ma2024real}. Thus, the insight behind risk-aware selective CP serves as the foundation for the communication-efficient strategy proposed in this study.

The second challenge concerns how to construct communication pathways in a scalable and dynamic traffic system. To the best of our knowledge, most existing CP studies are conducted within \textbf{Pre-arranged} communication zones and among predefined partner vehicles, as illustrated in Figure~\ref{fig:intro}(b). While pre-arranged communication enables stable point-to-point connections under idealized network assumptions—and even allows for global optimization of communication topologies and grouping strategies \citep{dong2022wave}—these setups are typically limited to experimental testbeds with a fixed number of specified agents. They fail to generalize to open-world traffic environments, where a small number of unfamiliar connected vehicles may encounter each other spontaneously across large spatial areas and at unpredictable times. This limitation highlights the urgent need for a decentralized and self-organized communication mechanism. To address this, we propose a \textbf{Spontaneous CP} framework that builds upon the selective CP principle. Each connected vehicle operates independently, broadcasting minimal information about its own perception coverage. Only when a risk-relevant blind zone is detected does it initiate a CP request. Neighboring vehicles, upon receiving the request, respond cooperatively if they are capable of contributing, as illustrated in Figure~\ref{fig:intro}(b). This mechanism leverages a key principle: connected vehicles can identify whether they can be assisted in blind-zone completion by evaluating other vehicles’ relative positions and their shared perception coverage. As a result, the handshake process is realized through lightweight broadcasting during regular operation and precise, selective information exchange triggered only when necessary.

To bridge the above two gaps, we propose a \textbf{Spontaneous Risk-Aware Selective Cooperative Perception (SRA-CP)} framework. It introduces a spontaneous collaboration mechanism composed of two modes: a routine mode, where vehicles continuously broadcast only their perception coverage maps; and a triggered mode, where a vehicle initiates CP only when it detects a risk-relevant blind zone, and neighboring connected agents are capable of assisting. Built upon this mechanism, we develop a \textbf{risk-aware hierarchical perception fusion model} that ensures efficient CP by adaptively prioritizing critical information within any available communication bandwidth. The model consists of four key components: a shared feature encoder, a risk-aware communication module, a dual-attention fusion network, and a multi-task decoder. This architecture enables vehicles to selectively fuse the most important perceptual features based on spatial occlusion and safety relevance under any given communication constraints. We evaluate the proposed framework on a public dataset by comparing it against three baselines: generic CP, a state-of-the-art selective CP method without risk awareness, and a no-CP setup, in terms of communication cost and Average Precision (AP) for key-object detection. Results demonstrate that SRA-CP achieves comparable or superior performance with significantly reduced communication overhead. Specifically, compared to generic CP, SRA-CP reduces the transmission volume to 80\% while incurring only a 0.1 drop in AP. Compared to the selective CP baseline without risk modeling, SRA-CP improves AP for critical objects by 15\% under the same communication budget, showcasing its potential for scalable deployment in large-scale, dynamic traffic environments.

\noindent The main contributions of this paper are as follows:
\begin{itemize}
    \item We address the bandwidth bottleneck in multi-agent CP by proposing a \textbf{risk-aware selective CP} strategy, which prioritizes the transmission of perceptual elements based on their impact on driving safety. This approach enables efficient use of limited communication resources under varying bandwidth constraints.
    
    \item We propose the \textbf{Spontaneous Risk-Aware Selective Cooperative Perception (SRA-CP)} framework, which supports dynamic, on-demand handshakes between agents without predefined regions or communication partners. This design enables scalable and low-cost CP in large-scale, real-world traffic environments with self-organizing connected agents.
    
    \item We validate the proposed framework on a public dataset and show that \textbf{SRA-CP} achieves comparable performance to generic CP while using only 20\% of the communication volume, with less than 1\% drop in AP. Compared to a cutting-edge selective CP baseline that does not consider driving risk, SRA-CP improves critical object detection accuracy by 15\%.
\end{itemize}

\section{Related work}

\subsection{Cooperative Perception (CP)}
CP allows multiple agents to share perceptual information to achieve a more complete understanding of their surroundings. This paradigm addresses key limitations of single-agent perception such as occlusion and limited sensing range \citep{chen2019fcooper,liu2020when2com}. There could be different downstream perception tasks to be fulfilled with CP, such as 3D object detection \citep{xu2023v2v4real,li2024multiv2x,xiang2024v2xreal,yu2023v2xseq}, lane detection \citep{jahn2024enhancing,elboukili2025v2v_lane_keeping_assist}, object tracking \citep{chiu2024dmstrack,zimmer2024tumtrafv2x,zhong2025cooptrack}. CP has been implemented through various fusion schemes, including early \citep{chen2019cooper,yang2025densityearly}, intermediate \citep{liu2020when2com,wang2020v2vnet,xu2022v2xvit}, and late fusion \citep{liu2024adaptive,sarlak2025extended_visibility}. 
Early fusion directly shares raw sensor data (e.g., LiDAR point clouds or images), aligns them in a common coordinate frame, and jointly processes the fused measurements through a single perception network. 
Intermediate fusion exchanges intermediate feature maps extracted by each agent’s backbone. These features are spatially aligned and combined before the detection head. 
Late fusion transmits only high-level perception outputs such as object boxes or tracks, which are then associated and merged at the cooperative layer, offering low communication cost at the expense of reduced ability to recover missed local detections.
 
Although these schemes achieve a certain trade-off between communication cost and perception fusion performance, the widespread transmission of perceptual information in multi-agent scenarios remains a significant challenge.

\subsection{Selective Information Sharing}
A major challenge in CP is reducing the communication burden while maintaining perception quality. Recent work such as Where2comm \citep{hu2022where2comm} has proposed to use spatial confidence maps to identify perceptually important regions and selectively transmit only the features from those areas. This strategy improves the perception accuracy under limited bandwidth by avoiding indiscriminate sharing of all information. There are also other CP works that share the same thoughts \citep{yang2023how2comm,liu2020when2com,yang2023what2comm}. However, perceptual accuracy alone is not a sufficient criterion in driving scenarios. In practice, much of the perceptual improvement may not contribute to driving decisions or safety \citep{ma2024real,vanbrummelen2018perception,pan2024reliability,gao2024vehicle}. For example, perceiving a distant vehicle with higher precision may not change the ego vehicle’s behavior. Thus, a key limitation of current selective strategies is the lack of risk-awareness. They fail to distinguish between information that is perceptually useful and information that is safety-critical. A risk-aware selection mechanism is needed to ensure that perception elements that are both unseen and pose potential safety risks are shared.

\subsection{Communication Paradigms for CP}

The communication paradigm for CP is also an important topic in CP. In most cases, prior studies assume a limited and fixed set of collaborators operating within a bounded area. These assumptions simplify the design of interaction protocols and enable direct coordination \citep{feng2018spatiotemporal,yu2019corridor}. Most existing CP frameworks also adopt such settings \citep{chen2019fcooper,liu2020when2com}. However, these conditions are difficult to satisfy in real-world traffic environments, where agents are numerous, highly dynamic, and distributed across large spatial regions. In addition, only a small fraction of encounters truly require cooperation, and participating agents are often unfamiliar with one another. These limitations highlight the need for a scalable, deployable, and self-organized CP communication paradigm. Recent work in agentic AI has begun to explore the notion of \emph{spontaneous cooperation} \citep{wu-etal-2024-shall,godhwani2025spontaneous,mirsky2022adhocteamwork}, where collaboration emerges dynamically based on local context and shared objectives. Building on this insight, our proposed SRA-CP framework leverages the property that connected vehicles can share their perception coverage, allowing other agents to identify opportunities to fulfill blind-zone completion needs. This enables the spontaneous formation of cooperation links without prior coordination or global knowledge.

\section{Problem Description}

\begin{figure}
    \centering
    \includegraphics[width=1\linewidth]{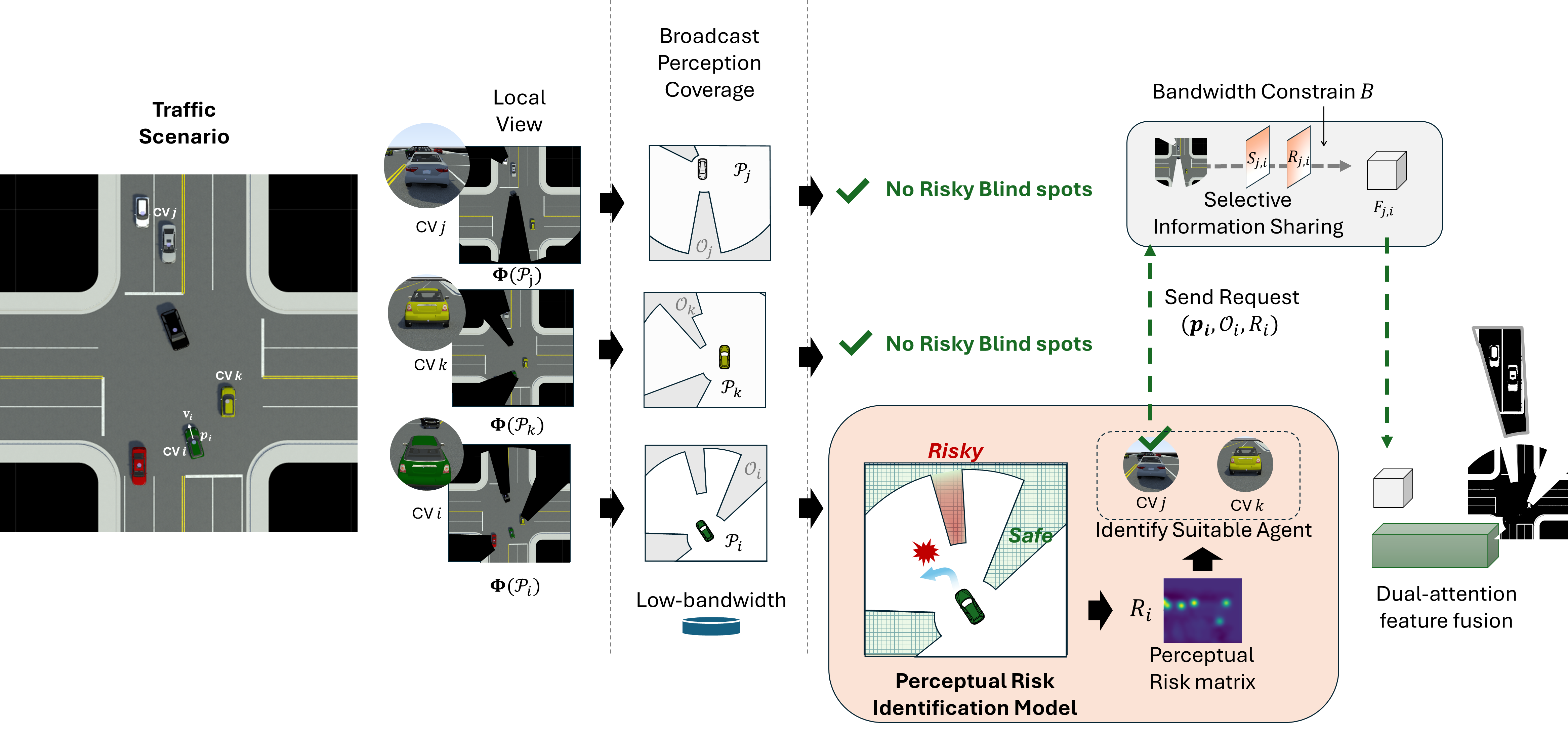}
    \caption{Spontaneous Risk-Aware Selective Cooperative Perception (SRA-CP)}
    \label{fig:problem}
\end{figure}

As illustrated in Figure~\ref{fig:problem}, we consider a dynamic road network with multiple Connected Vehicle (CV) agents indexed by $e,j,k$, in which $e$ denotes the ego vehicle. At a certain time $t$ (all variables hereafter are defined at time $t$ unless otherwise specified), each vehicle has a physical state represented by its position $\mathbf{p}_e= (x_e, y_e)$ and velocity $\mathbf{v}_e = (v^x_e, v^y_e)$, taking ego vehicle as an example. For a given connected agent $e$, its perception range from a bird's-eye view (e.g., the spatial coverage of LiDAR) is denoted by $\mathcal{P}_e$, with the corresponding sensed information $\boldsymbol{\Phi}(\mathcal{P}_e)$. The blind zone—areas not observable by the agent—is denoted as $\mathcal{O}_e $. In most cases, such blind zones have a negligible impact on driving safety, as illustrated by $\mathcal{O}_j$ and $\mathcal{O}_k$ in Figure~\ref{fig:problem}. However, in some cases, such as a vehicle approaching from the opposite direction within $\mathcal{O}_e$ that affects a left-turn decision, the blind zone can pose a significant risk. This study focuses on identifying such risky blind zones and selectively completing them via CP with limited communication bandwidth.

We design the SRA-CP framework in which the ego connected vehicle $e$ continuously broadcasts its own position $\mathbf{p}_e$, velocity $\mathbf{v}_e$, and perception coverage $\mathcal{P}_e$ (requiring only low bandwidth). It then receives broadcasted data from nearby vehicles within its communication zone $\mathcal{Z}_e$—a circular area of radius $l_c$—which includes the set of neighboring agents $\mathcal{W}_e$. Based on this shared information, the agent determines whether it has a risky blind zone that can be supplemented by any surrounding vehicle, and if so, initiates a spontaneous CP handshake and performs cooperative fusion. Specifically, the ego connected vehicle $e$ first evaluates the risk level of its perception blind zones using the proposed \textbf{perceptual risk identification} model, resulting in a perceptual risk matrix $\mathcal{R}_e$. If no risky blind zones are detected (e.g., as in the case of other CVs $j$ and $k$), the process at $t$ terminates. If risky blind zones are identified, the vehicle proceeds to select an appropriate target connected agent. Based on the shared perception coverage from neighboring agents (e.g., $\mathcal{P}_j$ and $\mathcal{P}_k$), the vehicle determines whether any connected agent can compensate for its occluded regions (e.g., agent $j$ in the illustrated case). Note that if no suitable connected agents are available to provide blind-zone compensation, the ego vehicle relies solely on its own onboard perception for decision-making—e.g., by stopping to continue observation. Once a candidate is selected, ego vehicle $e$ sends a CP request to agent $j$, including its current position $\mathbf{p}_e$, blind zone $\mathcal{O}_e$, and the computed risk matrix $\mathcal{R}_e$. Upon receiving the request, agent $j$ invokes the proposed \textbf{selective information sharing} model, which, under the given bandwidth constraint $B_{\mathrm{bytes}}$, selects and transmits the most informative features $F_{j,e}$ to supplement agent $e$'s perception. Finally, agent $e$ performs cooperative fusion via a \textbf{dual-attention feature fusion} model to integrate the received features and complete its understanding of the occluded region.

In this study, we focus on CP using LiDAR data, which provides accurate geometric structure, consistent performance under varying illumination, and reliable spatial measurements for dynamic traffic environments. These properties make LiDAR particularly suitable for blind-zone estimation and risk-aware perception. It is worth noting that the proposed framework is modality-agnostic. Although our implementation uses LiDAR as the primary sensing modality, the methodology can be extended to other perception inputs, such as video data.

\begin{figure}
    \centering
    \includegraphics[width=\textwidth]{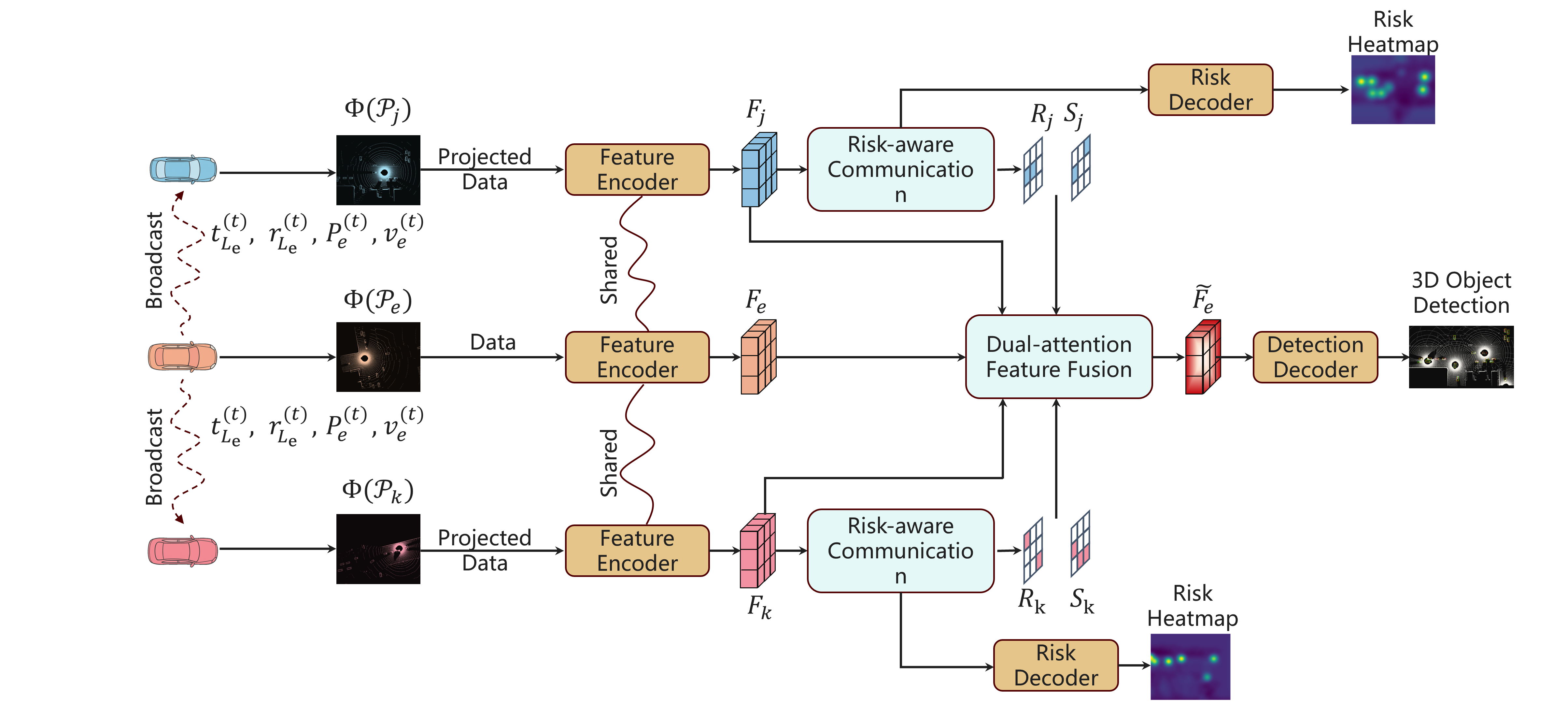}
    \caption{\textbf{End‑to‑end architecture of Selective Information Sharing and Fusion.}
    Each co‑operative vehicle $i \in \{e, k, j\}
$ projects its raw point cloud
    $\boldsymbol{\Phi}\!\left(\mathcal{P}_{i}\right)
$ to the ego Bird’s‑Eye‑View (BEV) frame and encodes it through a shared
    feature‑encoder, yielding $F_{i}$.
    \textbf{Risk‑aware communication} (Sec.~\ref{sec:risk_comm}) attaches two light‑weight
    masks—the spatial mask $S_{i}$ and the risk mask $R_{i}$—to the feature map
    and broadcasts only these three tensors, avoiding transmission of raw point clouds.
    The ego car receives the partner streams and performs
    \textbf{dual‑attention feature fusion} (Sec.~\ref{sec:dual_attention}):
    a safety‑focused selector prunes partner features with $(S_{i},R_{i})$ and a
    location‑wise multi‑head attention block aligns the surviving cells with the ego map
    $F_{e}$, producing $\tilde{F}_{e}$.
    Finally, two heads operate on $\tilde{F}_{e}$:
    (i) a Risk Decoder refines a dense risk heat‑map, and
    (ii) a Detection Decoder outputs 3‑D bounding boxes.
    }
    \label{fig:overall_arch}
\end{figure}

\subsection{Notation and Symbols}
\label{sec:notation}

{\small
\begin{longtable}{@{}llp{11cm}@{}}
\caption{Notation used throughout the paper (unified).}
\label{tab:notation}\\
\toprule
\textbf{Symbol} & \textbf{Type} & \textbf{Meaning / Unit} \\
\midrule
\endfirsthead

\multicolumn{3}{l}{\small\itshape Table~\ref{tab:notation} (continued).}\\
\toprule
\textbf{Symbol} & \textbf{Type} & \textbf{Meaning / Unit} \\
\midrule
\endhead

\midrule
\multicolumn{3}{r}{\small\itshape Continued on next page}\\
\endfoot

\bottomrule
\endlastfoot

\multicolumn{3}{l}{\emph{Sets, indices, regions}} \\
$t$ & scalar & Time index. \\
$i,j,k$ & index & Generic agent indices. \\
$e$ & index & Ego agent. \\
$\mathbf{p}_i=(x_i,y_i)$ & vector & 2D position of agent $i$ in BEV/ego frame (at time $t$, unless stated otherwise). \\
$\mathbf{v}_i=(v_i^x,v_i^y)$ & vector & 2D velocity of agent $i$. \\
$\mathcal{P}_i$ & set & Perception coverage (field of view, FoV) of agent $i$. \\
$\boldsymbol{\Phi}(\cdot)$ & map & Region $\to$ perceived sensory information; e.g., $\boldsymbol{\Phi}(\mathcal{P}_i)$ is the perception information of agent $i$. \\
$\mathcal{O}_i$ & set & Blind zone of agent $i$. \\
$\mathcal{Z}_{i}$ & set & Local communication region of agent $i$ (disk of radius $l_c$). \\
$l_c$ & scalar & Communication radius defining $\mathcal{Z}_{i}$ (m). \\
$\mathcal{W}_i$ & set & Neighboring connected agents within $\mathcal{Z}_{i}$. \\

\midrule
\multicolumn{3}{l}{\emph{SRA-CP protocol}} \\

$\rho_{i,j}$ & scalar & Pairwise collision risk score between $i$ and $j$. \\
$\tau_r$ & scalar & Threshold on $\rho$ to trigger risk-aware sharing. \\
$\mathcal{R}$ & matrix & Risk matrix collecting $\rho_{i,j}$. \\
$R_i,\ R_i^{(d)},\ R_i^{(s)},\ R_i^{n}$ & scalar & Object $i$'s total risk and distance/speed/intersection components. \\
$\hat{R}_i$ & scalar & Clipped/normalized risk of object $i$ in $[0,1]$. \\
$R_{\mathrm{gt}}$ & map & Ground-truth risk heatmap. \\
\(\mathcal{D}_e\) & set & Potentially dangerous agents for ego $e$, selected from $\mathcal{W}_e$ using $\mathcal{R}$. \\
\midrule
\multicolumn{3}{l}{Perceptual risk identification model}\\
$\mathbf{u}=(x,y)$ & vector & 2D BEV grid cell center in ego coordinates. \\
$\mathcal{G}_\mathbf{u}$ & grid & BEV grid cell centered at $\mathbf{u}=(x,y)$. \\
$o(\mathbf{u})\in[0,1]$ & field & Occupancy at BEV cell $\mathbf{u}$. \\
$\Pi_{\mathrm{BEV}}$ & op & 3D$\to$BEV projection operator. \\
$\kappa$, $\sigma(\cdot)$ & func & Smoothing kernel; squashing function for occupancy. \\
$\vartheta$ & angle & Ray azimuth (rad). \\
$r$ & scalar & Range of BEV grid cell (m). \\
$\mathbf{r}(s;\vartheta)$ & curve & Ray parameterization along azimuth $\vartheta$. \\
$T(\mathbf{u})$ & scalar & Line-of-sight transmittance to $\mathbf{u}$. \\
$\lambda$, $\Delta s$, $K$ & scalars & Beer–Lambert attenuation; step; number of samples along a ray. \\
$\chi_{\mathrm{fov}}(\mathbf{u})$ & gate & FoV gate $\{0,1\}$. \\
$P_{\mathrm{occ}}(\mathbf{u})$ & prob & Occlusion probability at $\mathbf{u}$. \\
$\tau_{\mathrm{occ}},\ K_t,\ \tau_t$ & scalars & Occlusion threshold; number of temporal frames; temporal consensus threshold. \\
$\mathcal{O}_{\mathrm{e}}(\mathbf{u}),\ \bar{\mathcal{O}}_{\mathrm{e}}$ & mask & Instantaneous and stabilized blind-zone masks. \\
$\mathbf{T}_{e\leftarrow w}$ & matrix & Rigid transform from world to ego frame. \\
$z$ & scalar & Vertical coordinate (height) in the ego frame. \\
\midrule
\multicolumn{3}{l}{\emph{Selective information sharing and fusion}} \\
$F_i\in\mathbb{R}^{C\times H\times W}$ & tensor & BEV feature map of agent $i$; $C$ channels, $H\!\times\!W$ the height/width of grid. \\
$C_{s,j}$, $C_{r,j}$ & map & Spatial- and risk-confidence maps on partner $j$. \\
$S_j,\ R_j\in\{0,1\}^{H\times W}$ & mask & Spatial / risk masks (binary). \\
$\tilde{F}_i$ & tensor & Masked feature patch to transmit from partner $i$. \\
$F_{j,e}$ & tensor & Partner $j$'s selected feature patch transmitted to ego $e$. \\
$f_{\mathrm{enc}},\ f_{\mathrm{dec}}$ & net & Shared encoder; multi-task decoder. \\
$(\hat{C},\hat{B},\hat{R})$ & out & Class scores, 3D boxes, refined risk heatmap. \\
$K_{\text{sel}}$ & scalar & Top-$K$ selected cells for transmission. \\
$g_{\mathrm{sp}}(\mathbf{u}),\ g_{\mathrm{risk}}(\mathbf{u})$ & score & Spatial/risk gains used for selection. \\
$g(\mathbf{u}),\ \alpha$ & score & Combined gain and its mixing weight $\alpha\in[0,1]$. \\
$P_e$ & path & Planned trajectory used by the risk head of ego agent $e$ . \\
\midrule
\multicolumn{3}{l}{\emph{Training objective and evaluation}} \\
$\mathcal{L}_{\text{total}}$ & loss & Total training loss. \\
$\mathcal{L}_{\text{det}}$ & loss & Detection loss. \\
$\lambda_{\text{risk}},\ \lambda_{\text{comm}}$ & weight & Weights for risk regression and communication penalty. \\
$\phi(\text{usage};\text{target})$ & penalty & Hinge-style penalty on over-usage of bytes. \\
$B_{\mathrm{bytes}}$ & bytes & Target per-link byte budget. \\
$h_{\mathrm{hdr}}$ & bytes & Header/metadata overhead per message. \\
$b_{\mathrm{idx}},\ b_{\mathrm{feat}},\ b_{\mathrm{cell}}$ & bytes & Bytes per cell index / per feature value / per cell ($b_{\mathrm{cell}}{=}b_{\mathrm{idx}}{+}C\,b_{\mathrm{feat}}$). \\
$U$ & bytes & Actual bytes used in a batch. \\
$B_{\mathrm{batch}},\ L_b$ & count & Batch size; number of agents in sample $b$. \\
$M^{(b)}_{l,i,j}\in\{0,1\}$ & mask & Binary selection mask for sample $b$. \\
$\mathrm{AP},\ \mathrm{3DAP}(\theta)$ & metric & AP and 3D AP at IoU threshold $\theta$. \\
$TP(\theta),\ FP(\theta),\ FN(\theta)$ & count & True/false positives and false negatives at $\theta$. \\
$\mathcal{I}_{\text{risk}}(\tau)$ & set & Subset filtered by risk threshold $\tau$ for Risk-AP. \\
$\bpk$ & metric & $\Delta$Risk-AP per KB ($\Delta\text{Risk-AP}/\text{KB}$). \\
\midrule
\multicolumn{3}{l}{\emph{Risk label generation}} \\
$\alpha_d, \alpha_s, \alpha_n$ & scalar
& Weights for distance-, speed-, and intersection-based risk components in the overall risk score $R$. \\
$\lambda_d, \lambda_n$ & scalar
& Decay rates for distance-based and intersection-based risk terms. \\
$m$ & index & index of intersections.\\
$\mathbf{q}_m$ & vector
& Center location of the $m$-th intersection. \\
$\mathcal{Q} $ & set
& Set of all intersection center locations. \\
$v_i$ & scalar
& Speed magnitude of object $i$. \\
$\epsilon$ & scalar
& Small positive constant to avoid division by zero in the speed-based risk normalization. \\

\end{longtable}
}

\section{Methodology}
As mentioned in the previous section, the proposed SRA-CP framework is designed to operate in two phases: during normal operation, each vehicle broadcasts basic perception coverage information with minimal bandwidth; when a risk-relevant blind zone is detected, it initiates a targeted CP link and transmits only the most critical information within the available communication bandwidth. The framework relies on a \textbf{perceptual risk identification model} to assess the risk level of blind zones. Upon identifying a suitable cooperative agent, the responder employs a \textbf{selective information sharing model} to determine which features to transmit under bandwidth constraints. The receiving agent then performs cooperative fusion using a \textbf{dual-attention feature fusion model} to produce an enhanced perception result. The following subsections detail each of these four key components.

\subsection{SRA-CP Protocol}

The core idea of the proposed SRA-CP protocol is as follows: at each time $t$, the ego vehicle $e$ periodically broadcasts a compact coverage map of $\mathcal{P}_e$ to all nearby agents $\mathcal{W}_e$ within $\mathcal{Z}_e$. This map summarizes which BEV cells are currently visible and which are likely occluded (Sec.~\ref{sec:blind_zone}), without exposing raw sensor data. Each neighbor does the same, enabling every agent to infer who can potentially compensate for its blind zones. When a risky blind area is detected, the ego triggers an on-demand handshake with one suitable partner and proceeds with selective sharing and fusion under the current byte budget.
In practice, a risk threshold $\tau_r$ determines whether the detected blind-zone risk warrants initiating the cooperative handshake.
Then, based on the received coverage maps, each agent constructs an inter-object risk matrix $\mathcal{R} = [\, \rho(e, i) \;|\; i \in \mathcal{W}_{\text{e}} \,]$ by evaluating pairwise risks. From this vector, the potentially dangerous set $\mathcal{D}_{\text{e}}$ is identified. If an agent $i \in \mathcal{D}_{\text{e}}$ also lies in $\mathcal{O}_{\text{e}}$, the partner transmits only the features covering that region to assist perception completion.

For example, as illustrated in the intersection scenario in Figure~\ref{fig:problem}, there are six vehicles, among which $i$, $j$, and $k$ are connected agents. In the first step, each connected agent broadcasts its local perception coverage $\mathcal{P}_i$, $\mathcal{P}_j$, and $\mathcal{P}_k$ to others. Since only coverage maps are shared—without detailed perception content—this step incurs negligible communication overhead.

Next, each agent performs an inter-object risk estimation over the observed objects in the scene and generates a risk vector $\mathcal{R}$ to estimate whether they need external information from other agents to help with their perception. In this scenario, agent $k$ and agent $j$ find no risky blind spot, so they do not need further external information from other agents. However, agent $i$ finds it can not see the potentially risky objects in the blind zone of the black vehicle, which is risky to its driving intention, and agent $j$ finds it can help the detection of agent i. Therefore, agent $j$ sends the information of the potentially risky zone to agent $i$ to complete its perception.

\subsection{Perceptual risk identification model}
\label{sec:blind_zone}
The \textbf{Perceptual Risk Identification Model} takes the individual perception $\boldsymbol{\Phi}(\mathcal{P}_i)$ as input and produces a risk matrix $\mathcal{R}_i$ over the blind zone $\mathcal{O}_i$, indicating the safety-critical importance of each location with respect to the ego vehicle's driving decisions.

SRA-CP requires a light-weight, geometry-based estimate of the ego vehicle's blind zones to prioritize compensation from partners. We adopt a BEV visibility model that is fast, rule-based, and admits a continuous formulation for analysis. Let $\mathcal{G}_\mathbf{u}$ denote a BEV grid with cell centers $\mathbf{u}=(x,y)$ in ego coordinates, z is the vertical coordinate of $\mathbf{u}$ in the ego frame used for 2.5D occupancy computation, ego pose $\mathbf{T}_{e\leftarrow w}$ (world\,$\to$\,ego), and a 2.5D occupancy field $o(\mathbf{u})\in[0,1]$ obtained from the LiDAR sweep $\boldsymbol{\Phi}\!\left(\mathcal{P}_{i}\right)$ by height-thresholding and kernel density aggregation:
\begin{equation}
o(\mathbf{u}) \,=\, \sigma\!\left( \max_{z\in[z_{\min},z_{\max}]}\; \kappa * \sum_{\mathbf{p}\in \mathcal{L}} \delta\big(\Pi_{\mathrm{BEV}}(\mathbf{T}_{e\leftarrow w}\,\mathbf{p})-(\mathbf{u},z)\big) \right),
\end{equation}
where $\Pi_{\mathrm{BEV}}$ projects 3D points to the BEV cell, $\kappa$ is a spatial smoothing kernel, and $\sigma$ is a squashing function ensuring $o\in[0,1]$ (e.g., $\sigma(a)=1-e^{-a}$). For a BEV direction $\vartheta=\mathrm{atan2}(y,x)$ and range $r=\lVert\mathbf{u}\rVert_2$, define the ray parameterization $\mathbf{r}(s;\vartheta)=s\,[\cos\vartheta,\sin\vartheta]^\top$, $s\in[0,r]$. The line-of-sight transmittance to $\mathbf{u}$ is modeled by a Beer–Lambert integral over occupancy:
\begin{equation}
T(\mathbf{u}) \,=\, \exp\!\left(-\int_{0}^{r} \lambda\, o\big(\mathbf{r}(s;\vartheta)\big)\, \mathrm{d}s \right),\quad \lambda>0,\label{eq:transmittance}
\end{equation}
with discrete approximation on grid steps $\Delta s$:
\begin{equation}
T(\mathbf{u}) \approx \exp\!\left(-\lambda\,\Delta s\, \sum_{k=0}^{K} o\big(\mathbf{r}(k\,\Delta s;\vartheta)\big) \right),\ K\,\Delta s\approx r.\label{eq:discT}
\end{equation}
Cells outside the sensor field-of-view (FOV) or range are treated as fully occluded by an FOV gate $\chi_{\mathrm{fov}}(\mathbf{u})\in\{0,1\}$; we define the occlusion probability and the binary blind-zone mask as
\begin{equation}
P_{\mathrm{occ}}(\mathbf{u}) \,=\, 1 - \chi_{\mathrm{fov}}(\mathbf{u})\,T(\mathbf{u}),\qquad
\mathcal{O}_{\mathrm{e}}(\mathbf{u}) \,=\, \mathbb{I}\big[P_{\mathrm{occ}}(\mathbf{u})\!>\!\tau_{\mathrm{occ}}\big],\label{eq:occ}
\end{equation}
with threshold $\tau_{\mathrm{occ}}\in(0,1)$. To reduce flicker, we temporally stabilize the mask by warping the last $K_t$ frames into the current ego frame using odometry and taking a robust union:
\begin{equation}
\bar{\mathcal{O}}_{\mathrm{e}}(\mathbf{u}) \,=\, \mathbb{I}\Bigg[\frac{1}{K_t}\sum_{t'=t-K_t+1}^{t} \mathcal{O}_{\mathrm{e}}^{(t')}\!\big(\mathbf{T}_{e\leftarrow e(t')}(\mathbf{u})\big)\, >\, \tau_t\Bigg].
\end{equation}
The mask $\bar{\mathcal{O}}_{\mathrm{e}}$ is used  as a compressed coverage summary and to increase selection gains in risky blind zones. Specifically, let $g_{\mathrm{sp}}(\mathbf{u})$ and $g_{\mathrm{risk}}(\mathbf{u})$ are spatial/risk scores (Sec.~\ref{sec:risk_comm}), the budgeted gain can be
\begin{equation}
g(\mathbf{u}) \,=\, \alpha\, g_{\mathrm{sp}}(\mathbf{u})\, g_{\mathrm{risk}}(\mathbf{u}) \; +\; (1-\alpha)\, \bar{\mathcal{O}}_{\mathrm{e}}(\mathbf{u})\, g_{\mathrm{risk}}(\mathbf{u}),\ \alpha\in[0,1],
\end{equation}
which prioritizes risky and occluded cells under a rate/byte budget.

\subsection{Selective Information Sharing and Fusion}
\label{sec:selective_fusion}
The \textbf{Selective Information Sharing and Fusion Model} describes the full pipeline from selecting the target agent for cooperation, to determining which features to share, and finally to integrating the received features on the ego vehicle. The overall framework is illustrated below.

\begin{figure}
  \centering
  \includegraphics[width=.55\linewidth]{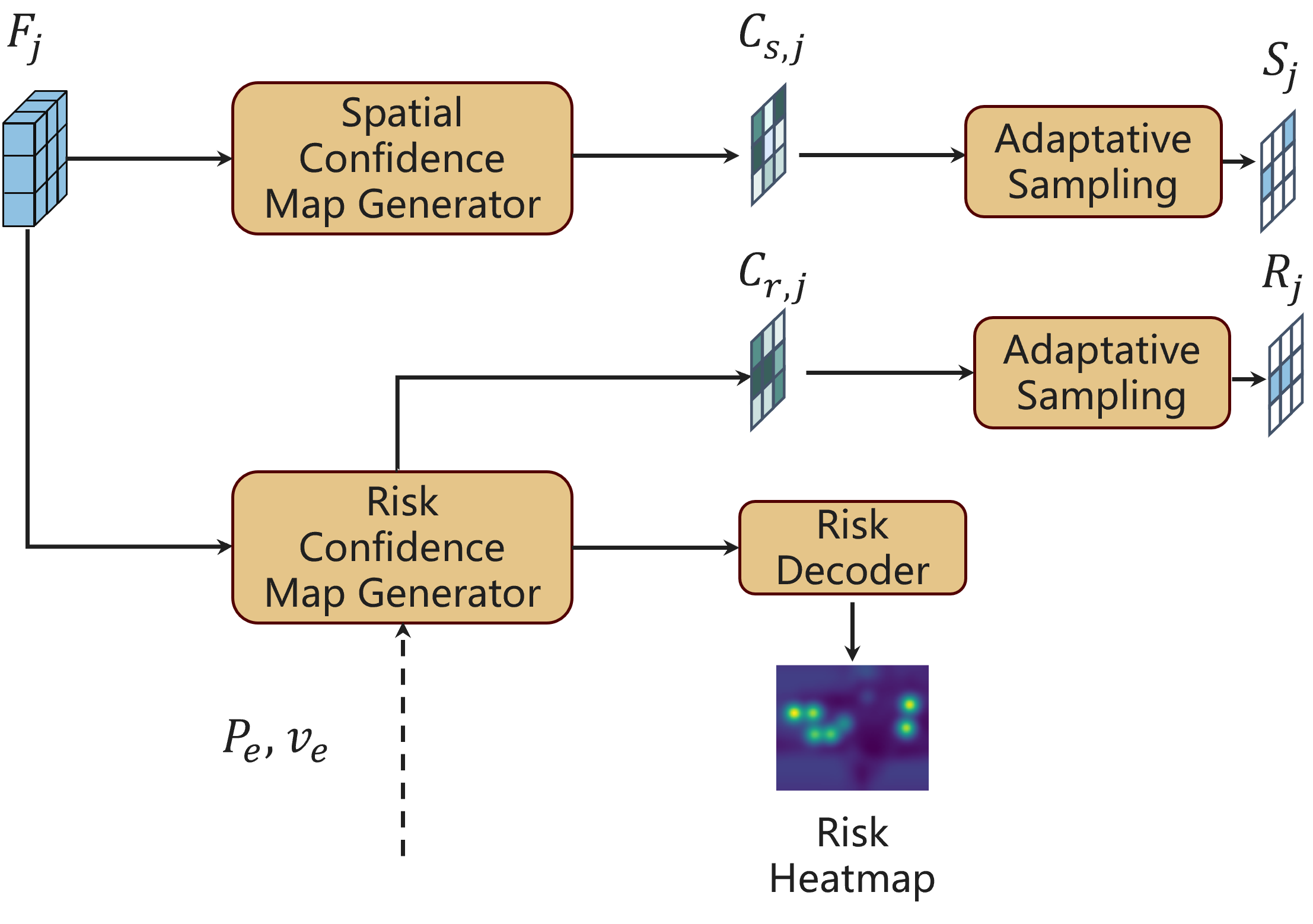}
  \caption{\textbf{Risk‑aware communication pipeline executed on each partner vehicle $j$.}
  The shared feature map $F_{j}$ is processed by two light‑weight heads:
  \emph{(i) Spatial‑confidence map generator} produces a spatial confidence map
  $C_{s,j}$ that highlights semantically important cells; an
  adaptive sampling module is used to select a sparse binary spatial mask based on scenario
  $S_{j}$ for transmission.
  \emph{(ii) Risk‑confidence map generator} uses $F_{j}$ together with the ego
  planned trajectory $P_{e}$ and speed $v_{e}$ to compute a risk map
  $C_{r,j}$.  Adaptive sampling converts it into a binary risk mask $R_{j}$.
  Both masks $\bigl(S_{j}, R_{j}\bigr)$ are sent to the ego vehicle, while a
  miniature Risk Decoder can optionally convert $C_{r,j}$ into a dense risk
  heat‑map for supervision training.
    }
  \label{fig:risk_comm}
\end{figure}

Selective Information Sharing and Fusion is the model layer that operationalizes the SRA-CP contract: given the neighbors and a per‑link budget, it learns what to communicate and how to fuse. Concretely, Selective Information Sharing and Fusion produces lightweight spatial and risk masks, sparsifies partner features under a given communication budget, and performs risk‑aware fusion on the ego agent using the fused ego map (Figure~\ref{fig:overall_arch}). This converts communication bandwidth into safety‑relevant detections by prioritizing risky × occluded regions. The pipeline has four building blocks:

\begin{enumerate}
    \item \textbf{Shared feature encoder} $f_{\mathrm{enc}}(\cdot)$ that transforms each
          LiDAR sweep $\boldsymbol{\Phi}\!\left(\mathcal{P}_{i}\right)$
 into a BEV feature tensor
          $F_i\!\in\!\mathbb{R}^{C\times H\times W}$;
    \item \textbf{Risk‑aware communication module}
          (Figure~\ref{fig:risk_comm}) that derives a spatial mask
          $S_i$ and a risk mask $R_i$ from $F_i$ and these masks will be used as a reference in the Dual-attention feature fusion process to decide which features should be shared;
    \item \textbf{Dual‑attention feature fusion} module
          (Figure~\ref{fig:dual_attention}) that selects the features $\{\tilde{F}_j\}_{j\neq e}$ to be shared to the ego vehicle based on the spatial mask
          $S_j$ and the risk mask $R_j$ and transmits
          $\{\tilde{F}_j\}_{j\neq e}$ to the ego vehicle and merges them with the ego feature map $F_e$
          and outputs the fused representation $\tilde{F}_e$ in the ego vehicle's coordinate system;
    \item \textbf{Multi‑task decoder} that predicts both 3D bounding boxes and a dense risk heat‑map.
\end{enumerate}

\vspace{.5em}
\subsubsection{Feature Encoding}
\label{sec:enc_risk}
During the training process each CV $ i$  encodes its LiDAR sweep $\boldsymbol{\Phi}(\mathcal{P}_i)$  with a shared PointPillar BEV encoder\cite{lang2019pointpillars} in the same structure, yielding $F_i\in\mathbb{R}^{C\times H\times W}$. Features are expressed in a common ego BEV frame using the known pairwise poses which is transmitted with the coverage map. The backbone within the same structure feeds two light heads to derive a spatial confidence map and a risk map. The spatial confidence map stores the confidence score of the features from the spatial perspective, which means which feature is spatially important for perception. And the risk confidence map stores the confidence score of the features from the traffic risk perspective, which means which feature is essential in terms of traffic importance. Both of these two confidence maps will guide the communication process to choose which features to communicate and the later fusion.

\vspace{.5em}
\subsubsection{Risk‑Aware Communication}
\label{sec:risk_comm}
The aim of this module is to reduces the communication bandwidth while preserving the balance of safety relevance and spatial relevance. Each partner summarizes where its features are informative (spatial saliency) and where they are safety‑critical for the ego (risk), then the CVs will combine the scores together to select which features are more important for the ego vehicle and they will send only the most important parts under a given budget.

\begin{figure}
  \centering
  \includegraphics[width=0.5\linewidth]{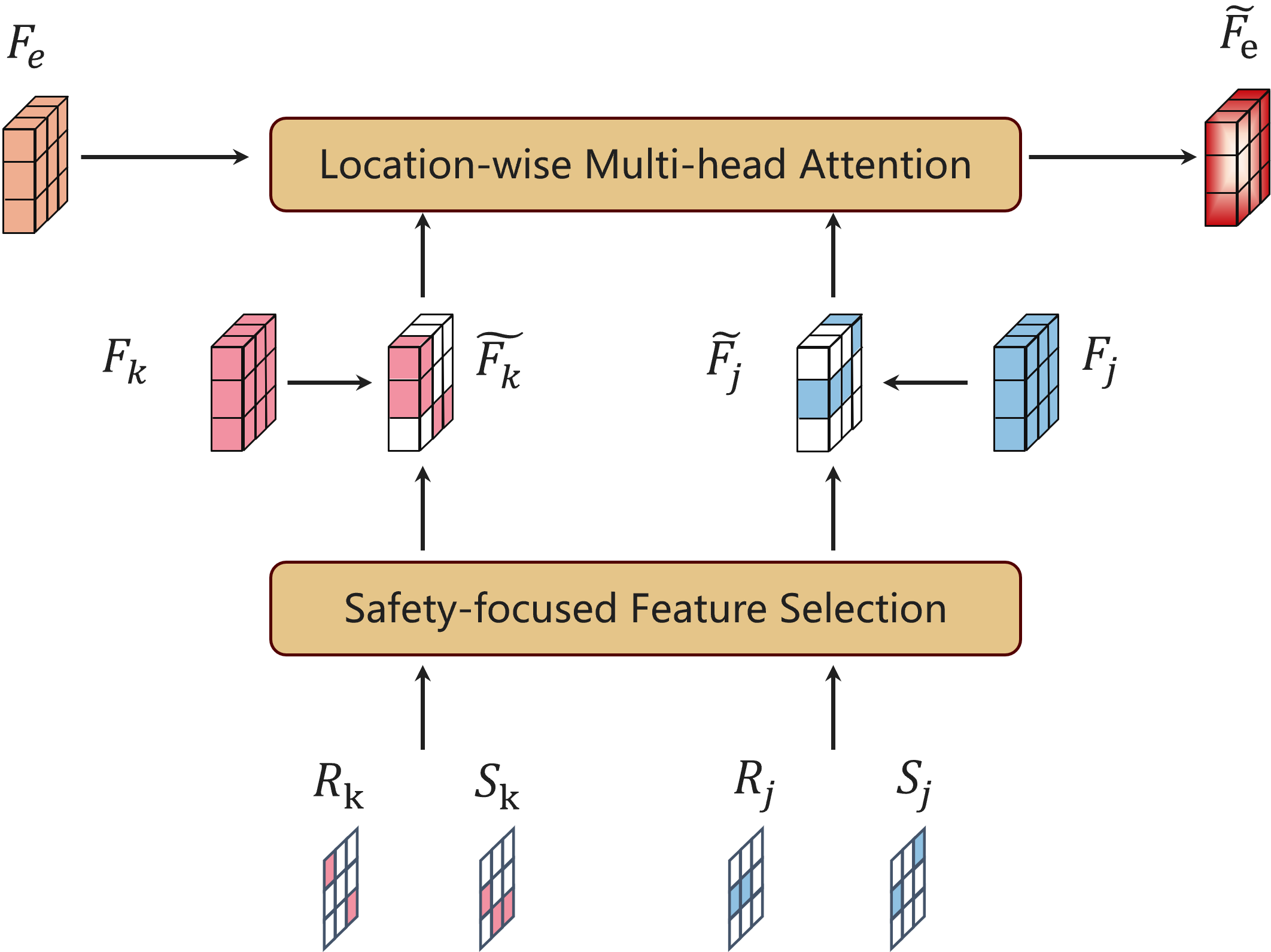}
  \caption{\textbf{Dual‑attention feature fusion.}
  Remote feature tensors $F_{k}$ and $F_{j}$ are first
  filtered by a Safety‑focused Feature Selection block that combines each partner’s
  spatial mask $S_{i}$ and risk mask $R_{i}$,
  yielding sparsified maps $\tilde{F}_{k}$ and $\tilde{F}_{j}$.
  The ego map $F_{e}$ and the sparsified partner maps are then fused by a
  location‑wise multi‑head attention module that performs per‑cell key–query
  interactions, producing an enriched representation $\tilde{F}_{e}$.
  This two‑stage design discards bandwidth‑hungry, low‑value regions
  before attention, so both communication and computation focus on areas that are
  simultaneously safety‑critical and semantically informative. During this process, only
  three low‑bandwidth tensors $(\tilde{F}_{j}, S_{j}, R_{j})$ leave the
  vehicle, preserving privacy and saving channel capacity.}
  \label{fig:dual_attention}
\end{figure}

On each partner $j$, two lightweight heads process $F_j$ (Figure~\ref{fig:risk_comm}):
\begin{itemize}
  \item Spatial‑confidence head outputs $C_{s,j}$ .
  \item Risk‑confidence head outputs $C_{r,j}$ .
\end{itemize}
Under a given communication budget, adaptive sampling will preform Top-K selection over non-ego grid cells based on their spatial and risk scores separately in the scene and then it will produces binary masks $S_j, R_j\in\{0,1\}^{H\times W}$. Under a given per‑link budget, SRA‑CP combines the two cues (union) and serializes only cells within the mask as in the safety-focus feature selection part in the Figure~\ref{fig:dual_attention}. In practice this masks the feature map: 
\[
\tilde{F}_j \,=\, F_j \odot \bigl(S_j \lor R_j\bigr),
\]
so only areas that are spatially salient and safety‑critical are transmitted. This concentrates communication on occluded or risky regions that matter for decision‑making, keeps privacy by avoiding raw points, and gracefully adapts to tighter budgets by shrinking the selected area.

\vspace{.5em}
\subsubsection{Dual‑Attention Feature Fusion}
\label{sec:dual_attention}

At the ego agent, the masked partner maps $\{\tilde{F}_j\}$ and the local map $F_e$ are fused in two stages (Figure~\ref{fig:dual_attention}). First, a safety‑focused selector re‑applies $\bigl(S_j,R_j\bigr)$ to suppress any residual clutter and enforce budget consistency. Second, We fuse ego and partner features in a location-wise manner, for each BEV cell $\mathbf{u}$, the ego feature provides the query, while only partners that selected this cell contribute keys and values. This yields an attention distribution over the relevant collaborators, ensuring that information is aggregated only where communication actually provided features. A residual update then produces the fused representation $\tilde{F}_e$.

To handle small spatial misalignment, the module can optionally attend within a local window around $\mathbf{u}$, but still restricts computation to cells indicated by partner selection. This keeps the complexity proportional to the number of communicated cells, making the fusion efficient under sparse CP.

This kind of design limits computation to a small set of safety‑relevant cells, improves alignment under occlusion, and avoids flooding the decoder with low‑value regions. When no partner data arrives, the module naturally falls back to the ego features without architectural changes.

\subsubsection{Budgeted Selection and Training Objective}

The fused tensor is decoded as
\(
(\hat{C},\hat{B},\hat{R}) = f_{\mathrm{dec}}\!\bigl(\tilde{F}_e\bigr),
\)
where $\hat{C}$ are class scores, $\hat{B}$ are 3‑D boxes, and
$\hat{R}$ is the refined risk heat‑map.

\paragraph{Budgeted selection.}
Given a budget per link, Selective Information Sharing and Fusion Model ranks non‑ego BEV cells by the gain $g(\mathbf{u})$ (Sec.~\ref{sec:blind_zone}) and selects the top $K_{\text{sel}}$ cells subject to the budget. Let the per‑cell byte cost be $b_{\mathrm{cell}}{=}b_{\mathrm{idx}}{+}C\,b_{\mathrm{feat}}$ and header overhead $h_{\mathrm{hdr}}$. For a byte budget $B_{\mathrm{bytes}}$, the capacity in cells is
\begin{equation}
K_{\mathrm{sel}} \;=\; \max\!\left(0,\; \left\lfloor \frac{B_{\mathrm{bytes}} - h_{\mathrm{hdr}}}{\;b_{\mathrm{cell}}\;} \right\rfloor \right),
\end{equation}
where, \(h_{\mathrm{hdr}}\) is a fixed header cost (bytes). 

\paragraph{Budget‑aware training.}
To make the bandwidth–accuracy trade‑off controllable at training time, we add a communication regularizer that penalizes over‑usage relative to a target budget; this does not change runtime budget, but shapes the model’s selection behavior. The total loss denoted by $\mathcal{L}_{\text{total}}$ is as follows:
\begin{equation}
\mathcal{L}_{\text{total}}
= \mathcal{L}_{\text{det}}
+ \lambda_{\text{risk}} \, \lVert \hat{R} - R_{\text{gt}} \rVert_2^2
+ \lambda_{\text{comm}} \, \phi(U; B_{\mathrm{bytes}}).
\end{equation}
where:
\begin{itemize}
    \item \textbf{Detection loss.}
    \(
    \mathcal{L}_{\text{det}} = \mathcal{L}_{\text{conf}} + \mathcal{L}_{\text{reg}}
    \)
    is the standard detection loss.  
    The classification term \(\mathcal{L}_{\text{conf}}\) is a focal loss with
    \(\alpha = 0.25\), \(\gamma = 2.0\),
    computed on BEV anchors and normalized by the number of positives.  
    The regression term \(\mathcal{L}_{\text{reg}}\) is a weighted Smooth-L1 loss over 7 box codes per anchor.

    \item \textbf{Risk regression.}  
    The risk loss is a mean-squared error between predicted and ground-truth BEV risk maps:
    \[
    \lVert \hat{R} - R_{\text{gt}} \rVert_2^2.
    \]
        \item \textbf{Communication over-usage penalty.}  
    The term \(\phi(U; {B_{\mathrm{bytes}}})=\max\big(0,\, U/B_{\mathrm{bytes}} - 1\big).\) penalizes only communication \emph{above} the target budget, aligning learned masks with the desired budget without changing the runtime protocol.

    The usage definitions can be calculated from:
    \begin{align}
    U = B_{batch} \cdot h_{\mathrm{hdr}}
    + \Bigg(\sum_{b=1}^{B_{batch}} \sum_{l=2}^{L_b}\sum_{i,j} M^{(b)}_{l,i,j}\Bigg)
    \cdot \big(b_{\mathrm{idx}} + C \cdot b_{\mathrm{feat}}\big),
    \end{align}
    where  
    \(B_{batch}\) is the batch size of this training,
    \(M^{(b)}_{l,i,j} \in \{0,1\}\) is the non-ego mask (for all the masks of ego is \(l{=}1\)),  
    \(h_{\mathrm{hdr}}\) is a fixed header bytes cost,  
    \(b_{\mathrm{idx}}\) is per-cell index bytes cost,  
    \(C\) is the channel dimension, and  
    \(b_{\mathrm{feat}}\) is bytes per feature value.

\end{itemize}

\section{Experimental Setup}

\subsection{Datasets}
\begin{figure}
  \centering
  \includegraphics[width=\linewidth]{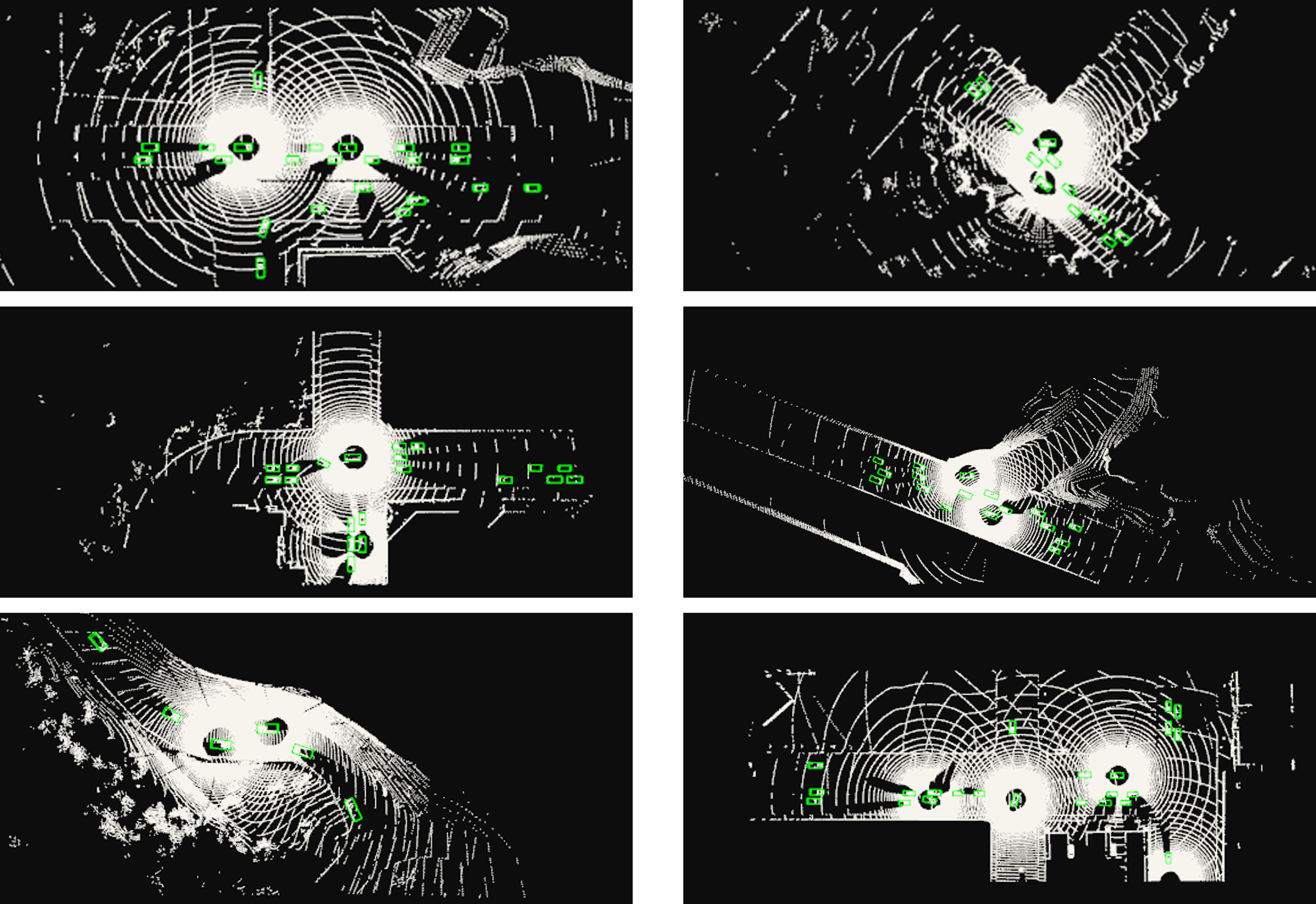}
  \caption{Representative exemplars from OPV2V illustrating scenarios that are inherently covered by the dataset and organized by us.}
  \label{fig:dataset_examples}
\end{figure}

We use the OPV2V dataset~\citep{xu2022opv2v} as the base dataset. OPV2V is a synthetic multi-vehicle CP benchmark generated by the OpenCDA co-simulation of SUMO~\citep{krajzewicz2012recent} and CARLA~\citep{dosovitskiy2017carla}. OPV2V contains 73 scenarios (average $\sim$25\,s) across multiple CARLA towns, where $2\!\sim\!7$ connected vehicles record 64-channel LiDAR from their own viewpoints. We follow the standard frame-level counts of 6765/1981/2170 for train/val/test, respectively. Importantly, OPV2V natively covers a diverse set of driving situations without any additional sampling from our side. The included situations comprise:
\begin{itemize}
    \item \textbf{Overtaking / Lane Change}: fast lateral maneuvers with transient occlusions.
    \item \textbf{Left-turn} and \textbf{Right-turn Intersections}: cross-traffic under partial observability (pedestrians/cyclists may emerge from blind zones).
    \item \textbf{On-ramp Merging}: gap selection and speed adjustment with strong temporal risk.
    \item \textbf{Unprotected Crossroads}: multiple agents with conflicting trajectories.
    \item \textbf{Head-on Encounters}: close-range opposing traffic forming highly critical regions.
    \item \textbf{Straight Driving (Low-risk Baseline)}: low-complexity scenes for calibration.
    \item \textbf{Multi-agent Cooperation}: $\geq$3 vehicles jointly negotiating maneuvers.
\end{itemize}
To illustrate why risk-aware cooperation is meaningful, we provide illustrative exemplars from OPV2V for the above situations in Figure~\ref{fig:dataset_examples}. These thumbnails are for visualization only and do not change the dataset composition.

To strengthen generalization and avoid leakage, we keep the natural scenario composition of OPV2V, but ensure that train/val/test have comparable proportions of each situation (e.g., intersections, merging, head-on). The unit of assignment is the entire scenario (all its frames stay in one split), preventing temporal leakage while reducing distributional drift between splits.

We control the number of agents per frame (2–7) by matching their histograms across splits within $\pm 5\%$. The qualitative exemplars in Figure~\ref{fig:dataset_examples} shows the different scenarios that are inherently covered by the dataset organized by us.

\subsection{Risk label generation}
To facilitate risk-aware CP using the OPV2V dataset~\citep{xu2022opv2v}, we generate risk annotations based on spatial, kinematic, and traffic-contextual information, further refined by expert domain knowledge—particularly in complex environments such as intersections. The final risk score for each object is computed as a weighted combination of three sub-components:
\begin{equation}
R_i = \alpha_d R_i^{(d)} + \alpha_s R_i^{(s)} + \alpha_m R_i^{(n)},
\end{equation}
where $R_i$ denotes the overall risk score for object $i$, and $R_i^{(d)}$, $R_i^{(s)}$, and $R_i^{(n)}$ correspond to distance-based, speed-based, and intersection-based risk scores, respectively. The weights $\alpha_d = 0.5$, $\alpha_s = 0.3$, and $\alpha_n = 0.2$ were selected based on empirical tuning and expert input.
\begin{itemize}
    \item \textbf{Distance-Based Risk:}  
Objects located closer to the ego vehicle are more likely to pose an immediate threat. We quantify this via an exponential decay function of the Euclidean distance:

\begin{equation}
R_i^{(d)} = \exp(-\lambda_d \cdot \| \mathbf{p}_i - \mathbf{p}_{e} \|_2),
\end{equation}
where $\mathbf{p}_i$ and $\mathbf{p}_{e}$ denote the positions of object $i$ and the ego vehicle, respectively. The parameter $\lambda_d$ controls the decay rate of risk with distance.

\item \textbf{Speed-Based Risk:}  
Rapidly approaching vehicles or those with high relative speed introduce dynamic hazards. We model this component as:
\begin{equation}
R_i^{(s)} = \frac{|v_i - v_{e}|}{\max_j |v_j - v_{e}| + \epsilon},
\end{equation}
where $v_i$ is the velocity of object $i$, $v_{e}$ is the ego vehicle’s speed, and $\epsilon$ is a small constant to avoid division by zero. This formulation emphasizes relative speed normalized across the scene.

\item \textbf{Intersection-Based Risk:}  
Intersections are inherently high-risk regions due to complex traffic flows, occlusion, signal compliance issues, and the presence of vulnerable road users. We begin by measuring proximity to intersections:
\begin{equation}
R_i^{(n)} = \exp\left(-\lambda_t \cdot \min_{\mathbf{q}_m \in \mathcal{Q}} \| \mathbf{p}_i - \mathbf{q}_m \|_2\right),
\end{equation}
where $\mathcal{Q} = \{\mathbf{q}_m\}$ denotes known intersection coordinates and $\lambda_m$ adjusts the decay with distance to intersections.

\end{itemize}

\paragraph{Normalization:}  
Finally, we clip the combined risk score to the range $[0, 1]$ for stable learning:
\begin{equation}
\hat{R}_i = \min(1, \max(0, R_i)).
\end{equation}

\subsection{Baselines}
To contextualize the standard AP results, we compare the following baselines under the same backbone, BEV grid, IoU thresholds, synchronization window, and quantization:
\begin{itemize}
  \item \textbf{Where2Comm (Spatial-only baseline) \citep{hu2022where2comm}.}
  A representative spatial-communication method that learns where to communicate 
  based solely on spatial saliency without explicit risk or task-aware weighting. 
  Each agent predicts a binary mask indicating informative BEV cells, 
  and only those regions are transmitted for feature fusion.  This baseline captures the benefit of geometry-aware but task-agnostic cooperation.
  
  \item \textbf{Upper Bound (fully connected).} 
  Fully connected communication with no budget, transmitting all partner features for fusion; serves as a performance ceiling.
  
  \item \textbf{Lower Bound (single-agent).} 
  No cooperative communication. It 
  measures the capability of the ego-only detector.
  
  \item \textbf{Fixed-Neighbor (equal-budget).} 
    The total communication budget is equally divided among all non-ego neighbors. Within each neighbor, features (e.g., grid cells or point clusters) are uniformly sampled at random. This baseline isolates the effect of adaptive link-wise budget allocation from uniform distribution.
  
  \item \textbf{Random-Cell.} 
    A global uniform sampler randomly selects exactly $K$ feature cells from all non-ego agents, regardless of their spatial location or risk relevance. This baseline evaluates the effectiveness of our selective content transmission compared to random feature selection under the same bandwidth constraint.
\end{itemize}

\subsection{Implementation details}
\textbf{Model and feature encoding.}
We adopt a PointPillar BEV backbone. The PillarVFE uses 64 channels. The BEV backbone has 3/5/8 blocks with filters 64/128/256 and deconvs of 128 channels; a shrink header downsamples to 256 channels for heads. We attach three lightweight heads: classification (per cell 2 anchors), regression (7 parameters per anchor), and a risk head (1 per anchor) to produce dense risk heatmaps.

\textbf{Voxel/grid and anchors.}
Voxel size is $0.4\times0.4\times4$\,m with LiDAR range $[-140.8,-38.4,-3,\;140.8,\,38.4,\,1]$\,m. The BEV grid is $H{=}192$, $W{=}704$ (feature stride 4). Anchors follow $(l,w,h){=}(3.9,1.6,1.56)$ with yaw $\{0^\circ,90^\circ\}$; $NMS=0.15$, and the positive, negative thresholds are $0.6$ and $0.45$ separately.

\textbf{Training setup.}
Optimizer: We select Adam with leanring rate=2\,$\times$\,10$^{-4}$ as the optimizer and the selection of weight\_decay is 0.01 and \texttt{eps}=1e-10. For the learning rate schedule, we set cosine annealing for 50 epochs with 10-epoch warmup (with the warmup learning rate=2\,$\times$\,10$^{-5}$, and the minimal learning rate=5\,$\times$\,10$^{-6}$). During the training of all the models, we set the batch size as 8. In terms of connecting agent numbers, we cut the number of agents up to 5 CAVs. Data augmentation includes x-axis flip, random rotation ($\pm45^\circ$), and scaling (0.95–1.05). Voxelization caps are 32 points/voxel, with train/test voxel maxima as 32k/70k separately.

\textbf{Inference and post-processing.}
We decode detection and risk heatmaps after fusion. Evaluation uses IoU $\in\{0.3,0.5,0.7\}$; risk-aware AP uses $\tau\in\{0.2,0.3,0.4\}$. We log per-frame communication rate and bytes for the report.

\subsection{Evaluation protocols and metrics.}
We use 3D Average Precision (3DAP) to assess object detection performance. Given a detection is considered correct if the Intersection over Union (IoU) between the predicted and ground-truth 3D bounding box exceeds a threshold $\theta$, the AP is computed based on the precision-recall curve.

We report 3DAP under three IoU thresholds:
\[
\theta \in \{0.3, 0.5, 0.7\},
\]
corresponding to different levels of localization strictness.

Let $TP(\theta)$, $FP(\theta)$, and $FN(\theta)$ be the number of true positives, false positives, and false negatives under threshold $\theta$, respectively. Precision and recall are defined as:
\begin{equation}
\text{Precision}(\theta) = \frac{TP(\theta)}{TP(\theta) + FP(\theta)}, \quad
\text{Recall}(\theta) = \frac{TP(\theta)}{TP(\theta) + FN(\theta)}.
\end{equation}

3DAP is then computed as:
\begin{equation}
\text{3DAP}(\theta) = \int_{0}^{1} \text{Precision}(\theta, r) \, dr,
\end{equation}
where $\text{Precision}(\theta, r)$ is interpolated at recall level $r$.

To assess the influence of risk understanding on perception, we compute 3DAP selectively over high-risk regions determined by thresholding the predicted risk map.

Let $\mathcal{I}_{risk}(\tau) = \{i \, | \, \hat{R}_i > \tau\}$ be the set of objects or regions identified as risky with a risk threshold $\tau$. We evaluate detection performance on this subset, denoted as $\text{3DAP}_{risk}(\theta, \tau)$:
\begin{equation}
\text{3DAP}_{risk}(\theta, \tau) = \text{3DAP evaluated on } \mathcal{I}_{risk}(\tau), \text{ with IoU threshold } \theta.
\end{equation}

We report results for:
\[
\theta \in \{0.3, 0.5, 0.7\}, \quad \tau \in \{0.2, 0.3, 0.4\}.
\]

This metric captures how well the model perceives objects in scenarios that are potentially dangerous or require immediate attention, reflecting the synergy between risk assessment and spatial awareness.

We evaluate under two protocols:
\begin{itemize}
\item \textbf{P1: Fixed-bandwidth.} Per-link budget $B_{\mathrm{bytes}}\in\{0.5,0.7,1,2,3,5,10\}$\,KB/frame. Each method is tuned per $B_{\mathrm{bytes}}$; we report Risk-AP, $\bpk$, and data transmission volume. This stresses efficiency at scarce bandwidth.
\item \textbf{P2: Fixed-performance.} Given a target Risk-AP (e.g., $\ge X$), we report minimal bytes and latency to reach it. This answers how much bandwidth is necessary for a safety line.
\end{itemize}

\section{Results and Discussion}

\subsection{Main Results (Standard AP)}
We report standard detection AP at IoU 0.3/0.5/0.7 denoted as AP30, AP50 and AP70 separately, across baselines and our method. See Table~\ref{tab:ap_comparison} for overall comparison. It should be noted that the communication budget of Where2comm, Fixed-Neighbor, Random-Cell and ours are 20\% of the fully connected situation like the settings of the method Upper Bound. And there is no communication in Lower Bound method.
As shown in the table, our model achieves consistently competitive performance across all IoU thresholds, with only marginal differences compared to the strongest baselines, while remaining close to the upper bound. This indicates that both our approach and the baseline methods are able to effectively leverage the advantages of CP.

\subsection{Risk-Aware Evaluation}
We further evaluate Risk-Aware AP by filtering ground-truths above risk thresholds $\tau \in {\{0.2, 0.3, 0.4\}}$. Results are summarized in Tables~\ref{tab:ap_comparison_risk_all}.
Compared with the overall AP results, where our model and the baselines perform similarly in Table~\ref{tab:ap_comparison}, the risk-aware evaluation reveals a clearer distinction. As shown in Table~\ref{tab:ap_comparison_risk_all}, our method consistently outperforms the baselines across all IoU thresholds, especially under higher risk conditions ($\tau=0.3, 0.4$). The performance of our model remains close to the upper bound while the spatial-only baseline drops significantly as risk increases. This demonstrates that our design better preserves detection robustness when encountering high-risk or safety-critical objects, validating the effectiveness of the communication protocol and SRA-CP coordination. In other words, although both methods achieve comparable aggregate perception accuracy, our framework exhibits stronger risk sensitivity and resilience, which are essential for safety-oriented CP. We additionally visualize risk-aware example heatmaps (Sec.~\ref{visual}).

\subsection{P1: Pareto efficiency under fixed bandwidth}
\begin{figure}
    \centering
    \begin{subfigure}[b]{0.32\textwidth}
        \includegraphics[width=\textwidth]{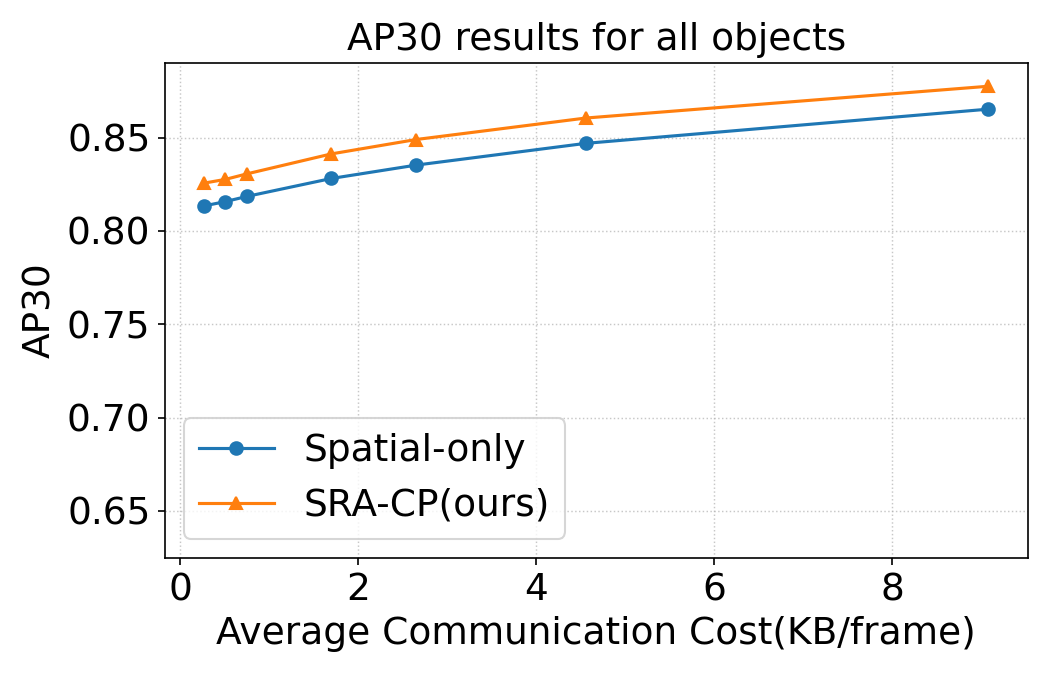}
        
        \label{fig:sub1}
    \end{subfigure}
    \begin{subfigure}[b]{0.32\textwidth}
        \includegraphics[width=\textwidth]{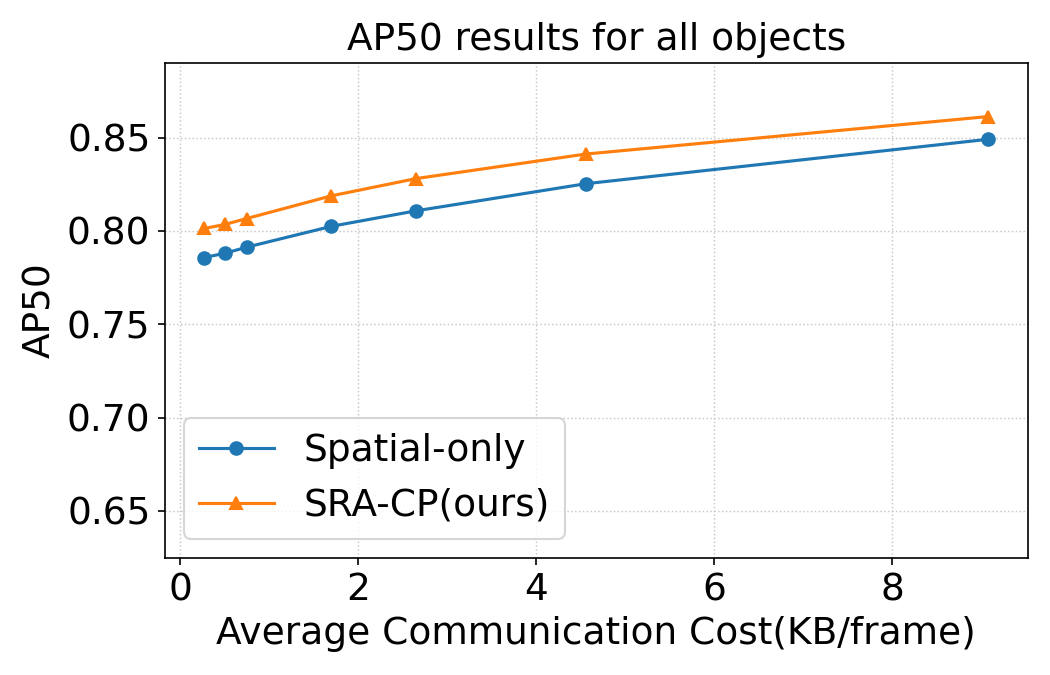}

        \label{fig:sub2}
    \end{subfigure}
    \begin{subfigure}[b]{0.32\textwidth}
        \includegraphics[width=\textwidth]{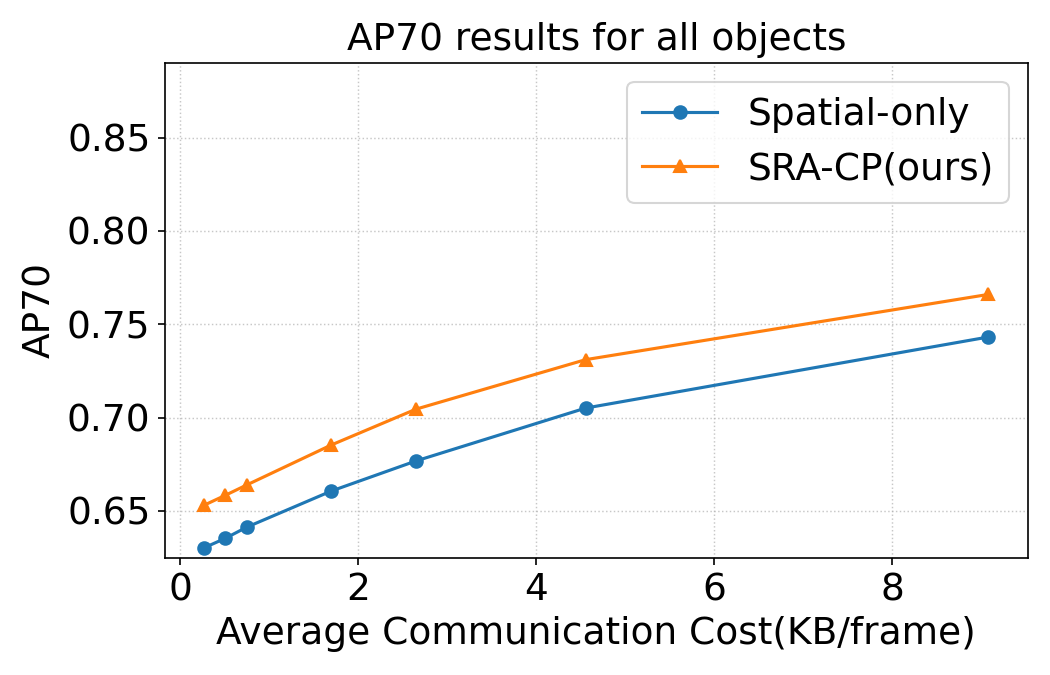}
    
        \label{fig:sub3}
    \end{subfigure}
    \caption{Comparison of perception accuracy (AP30, AP50, AP70) under varying communication costs (KB/frame) across all objects.}

    \label{fig:bytes_ap}
\end{figure}

\begin{figure}
    \centering
  
    \begin{subfigure}[b]{0.32\textwidth}
        \includegraphics[width=\textwidth]{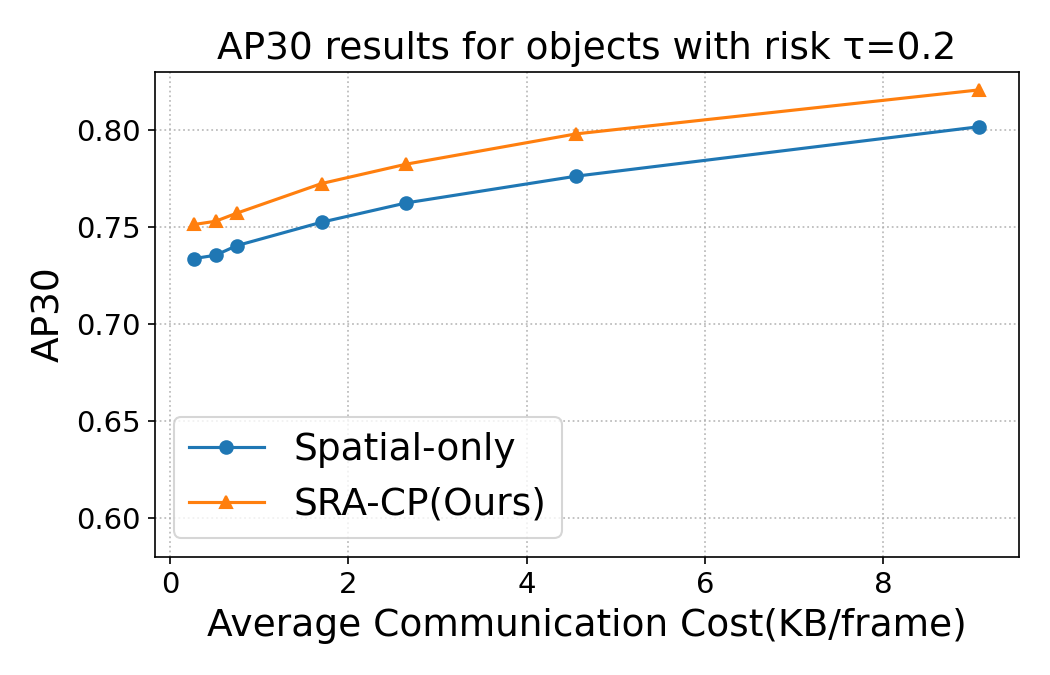}
        
    \end{subfigure}
    \begin{subfigure}[b]{0.32\textwidth}
        \includegraphics[width=\textwidth]{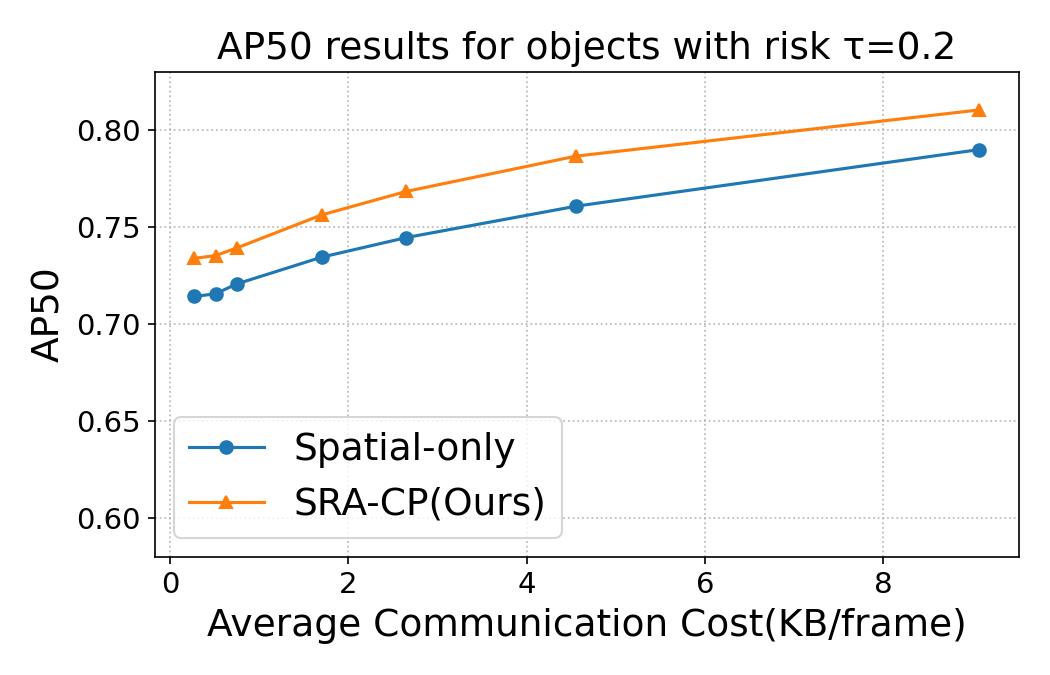}
        
    \end{subfigure}
    \begin{subfigure}[b]{0.32\textwidth}
        \includegraphics[width=\textwidth]{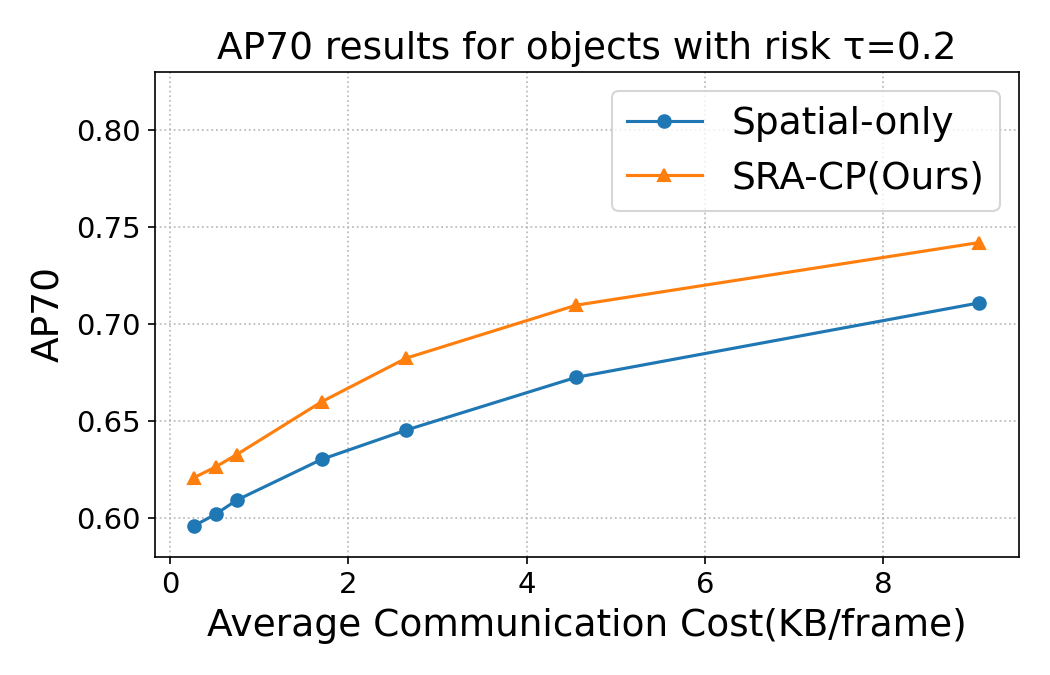}
       
    \end{subfigure}
    \\[1ex]

    \begin{subfigure}[b]{0.32\textwidth}
        \includegraphics[width=\textwidth]{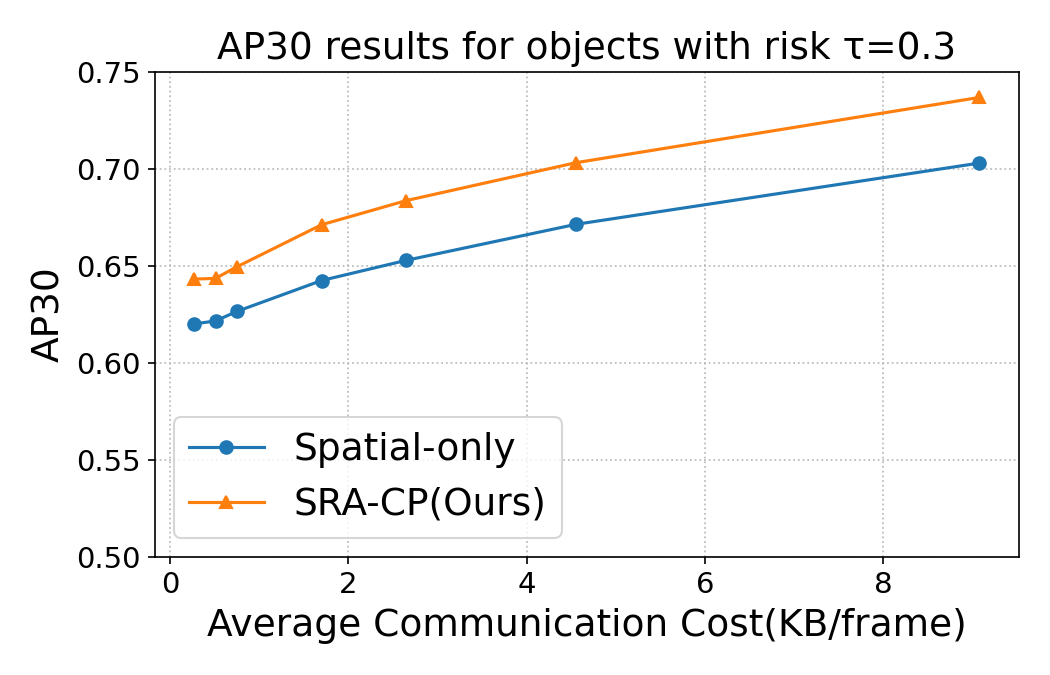}
       
    \end{subfigure}
    \begin{subfigure}[b]{0.32\textwidth}
        \includegraphics[width=\textwidth]{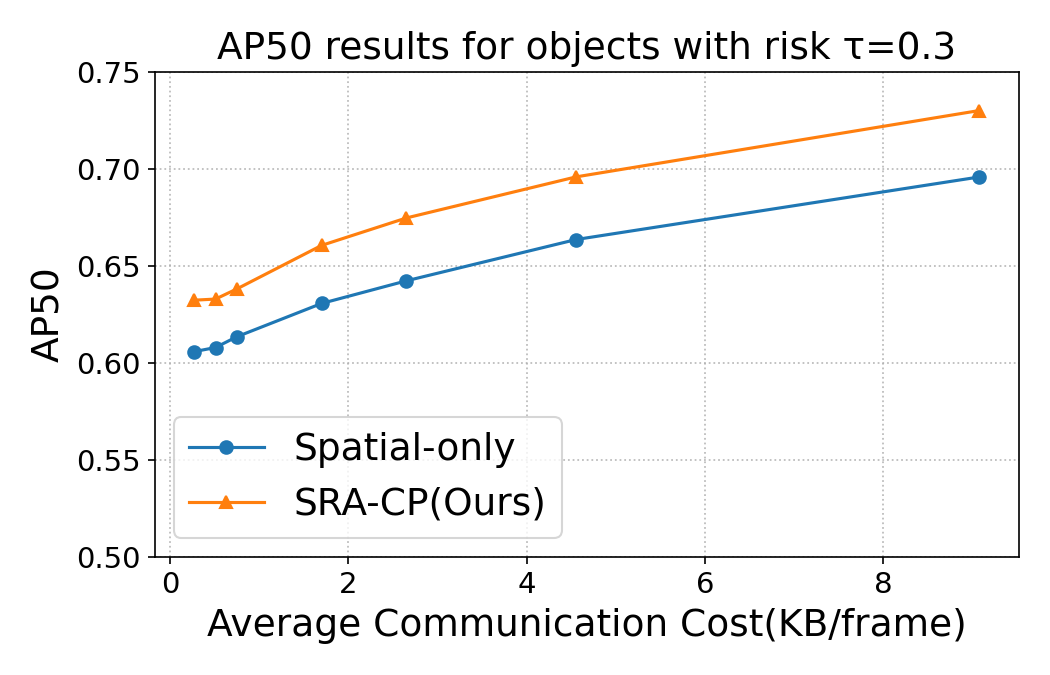}
        
    \end{subfigure}
    \begin{subfigure}[b]{0.32\textwidth}
        \includegraphics[width=\textwidth]{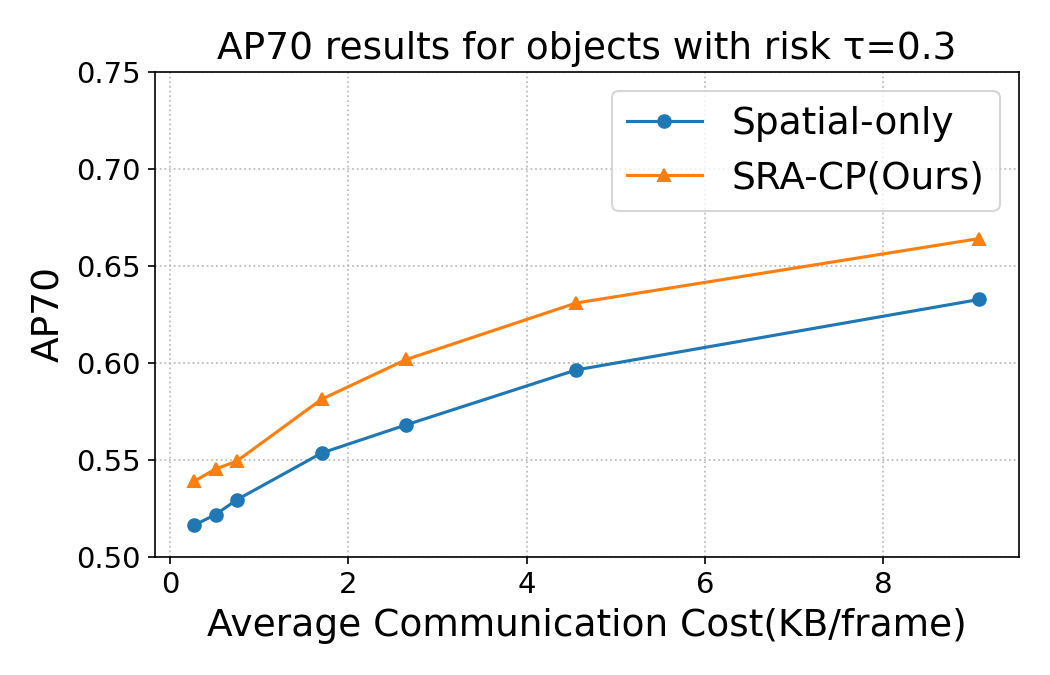}
       
    \end{subfigure}
    \\[1ex]

    \begin{subfigure}[b]{0.32\textwidth}
        \includegraphics[width=\textwidth]{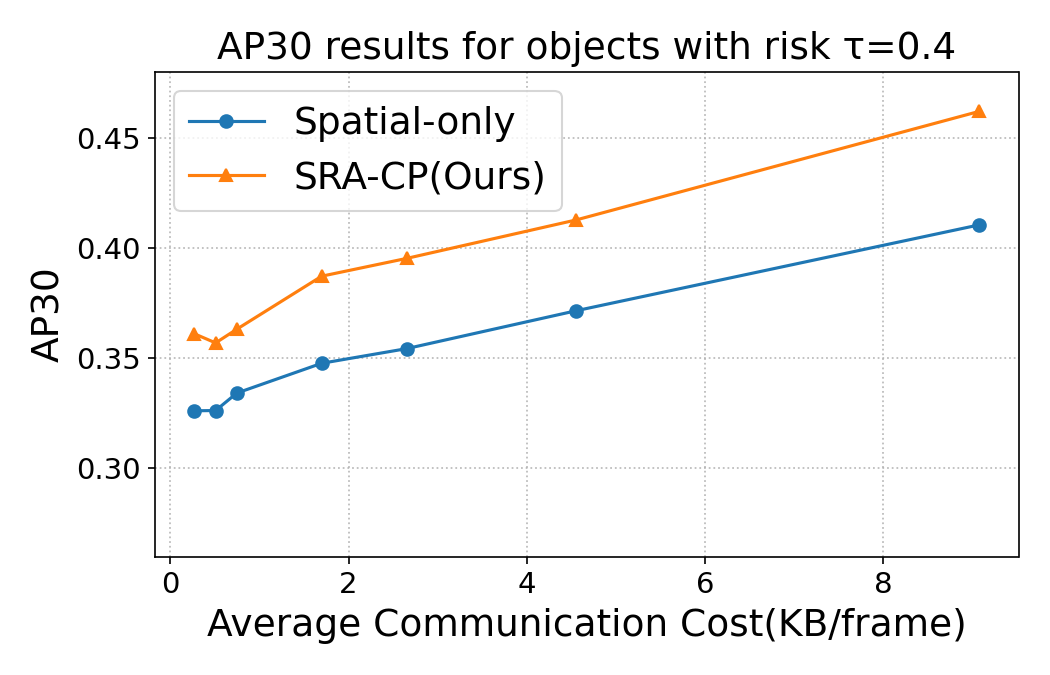}
        \label{fig:3a}
    \end{subfigure}
    \begin{subfigure}[b]{0.32\textwidth}
        \includegraphics[width=\textwidth]{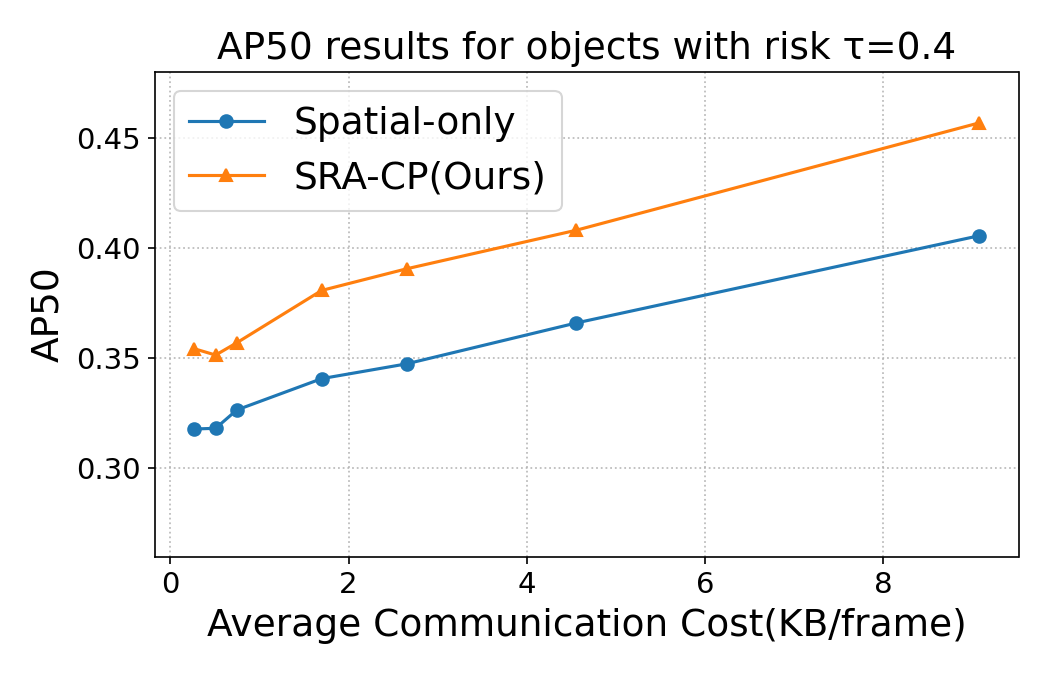}
        \label{fig:3b}
    \end{subfigure}
    \begin{subfigure}[b]{0.32\textwidth}
        \includegraphics[width=\textwidth]{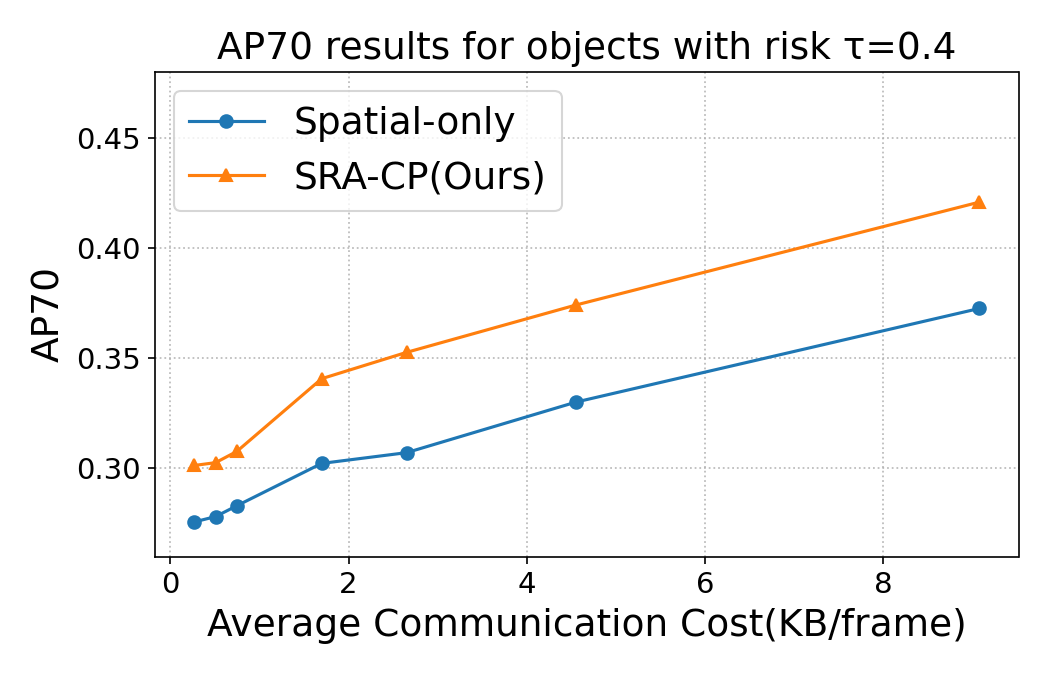}
        \label{fig:3c}
    \end{subfigure}
    \caption{Comparison of perception accuracy (AP@30, AP@50, AP@70) under varying communication costs (KB/frame) for objects with different risk levels, defined by risk thresholds $\tau \in \{0.2, 0.3, 0.4\}$.}
    \label{fig:pareto_risk}
\end{figure}

To further examine model performance under resource-constrained conditions, we plot AP30/AP50/AP70 vs. communication cost (KB/frame) in Figure~\ref{fig:bytes_ap} and Risk-AP30/AP50/AP70 vs. communication cost (KB/frame) in Figure~\ref{fig:pareto_risk}. These plots provide a quantitative view of how perception accuracy scales with bandwidth usage.

Across the 0.5–10 KB/frame regime, our proposed SRA-CP configuration consistently dominates the Pareto frontier, achieving higher safety-aware gains per byte compared to baseline methods. For example, at 5 KB/frame, our approach yields approximately +4.7\% Risk-AP50 improvement over the baseline while maintaining comparable communication overhead. This demonstrates that the method's communication sparsification and the inference fusion jointly enable efficient and safety-preserving cooperation.

\subsection{P2: Minimal cost to reach a safety line}
\begin{figure}
    \centering
   
    \begin{subfigure}[b]{0.32\textwidth}
        \includegraphics[width=\textwidth]{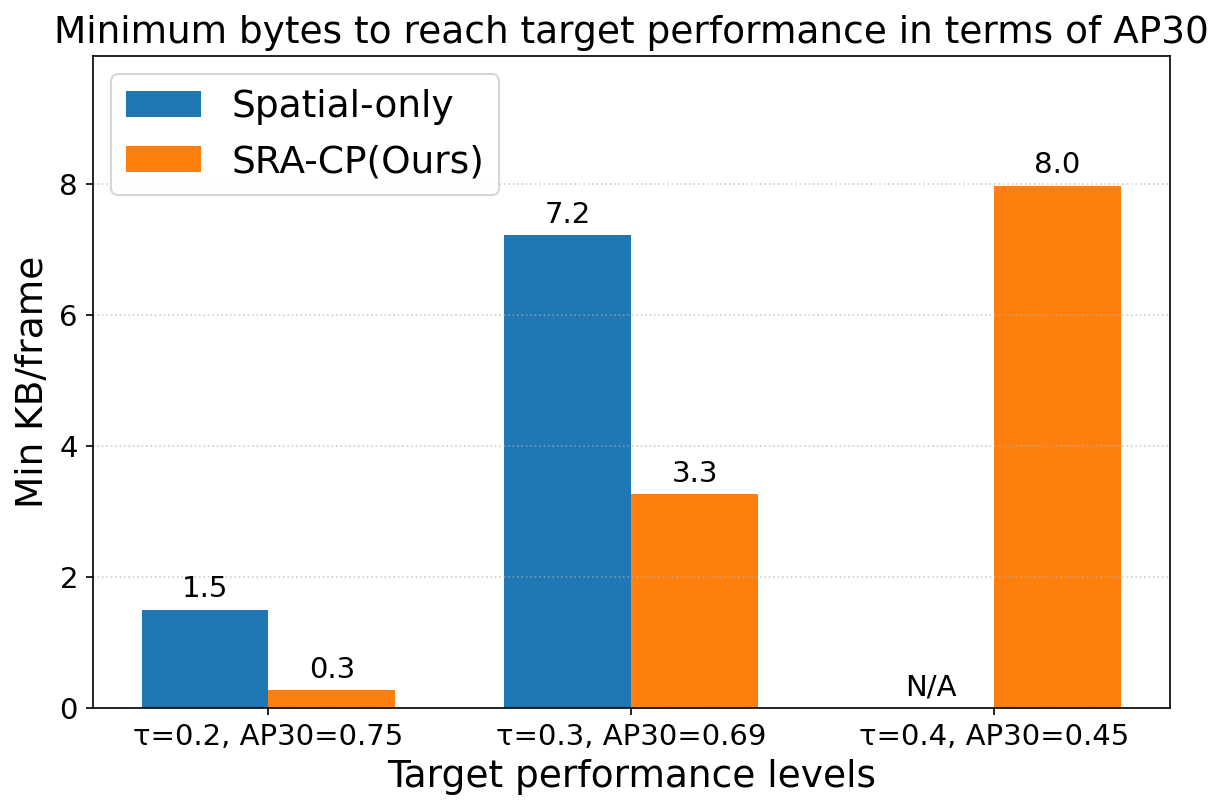}

    \end{subfigure}
    \begin{subfigure}[b]{0.32\textwidth}
        \includegraphics[width=\textwidth]{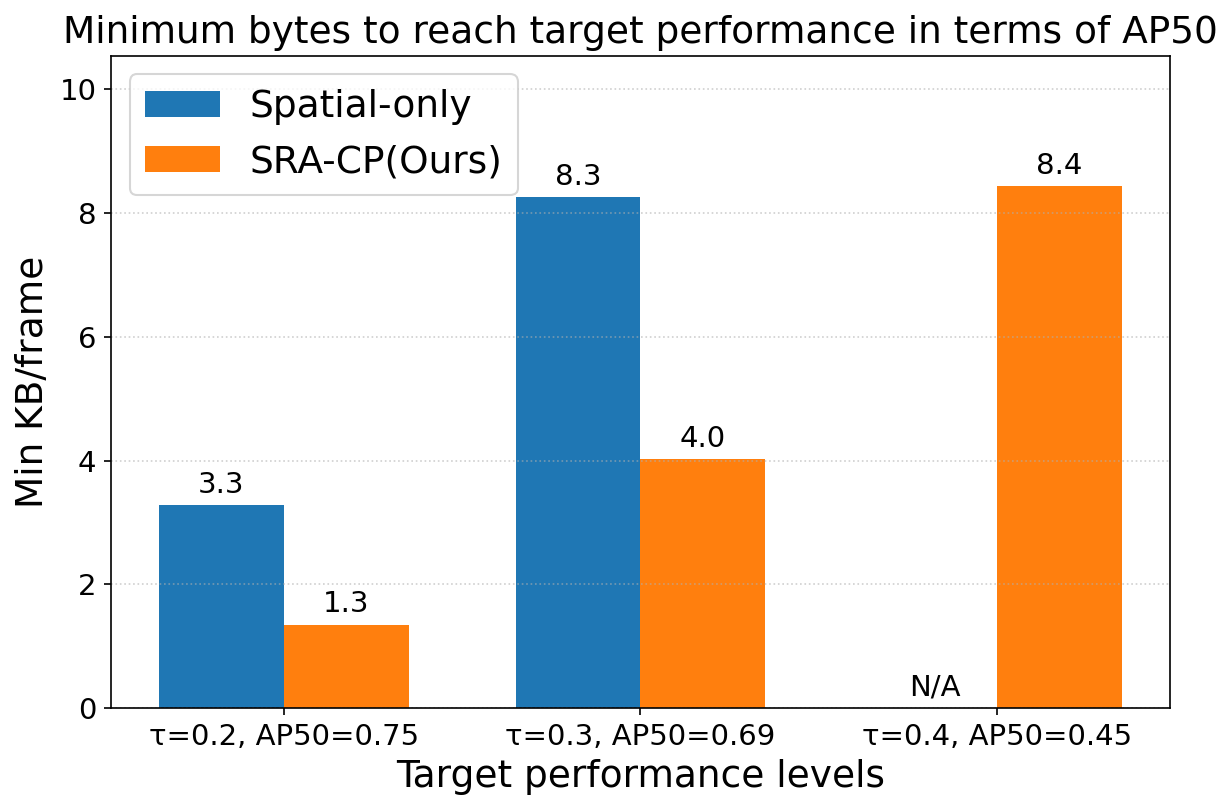}
        
    \end{subfigure}
    \begin{subfigure}[b]{0.32\textwidth}
        \includegraphics[width=\textwidth]{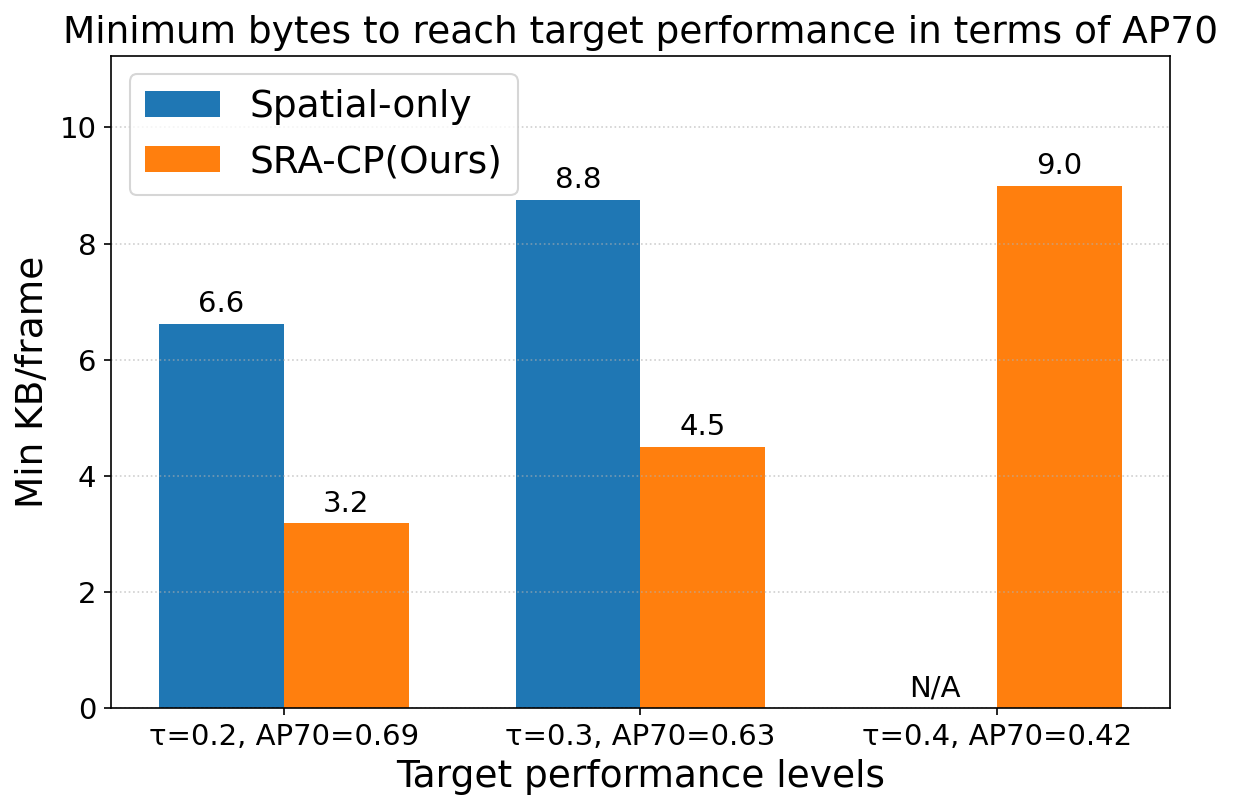}

    \end{subfigure}
    \\[1ex]

    \begin{subfigure}[b]{0.32\textwidth}
        \includegraphics[width=\textwidth]{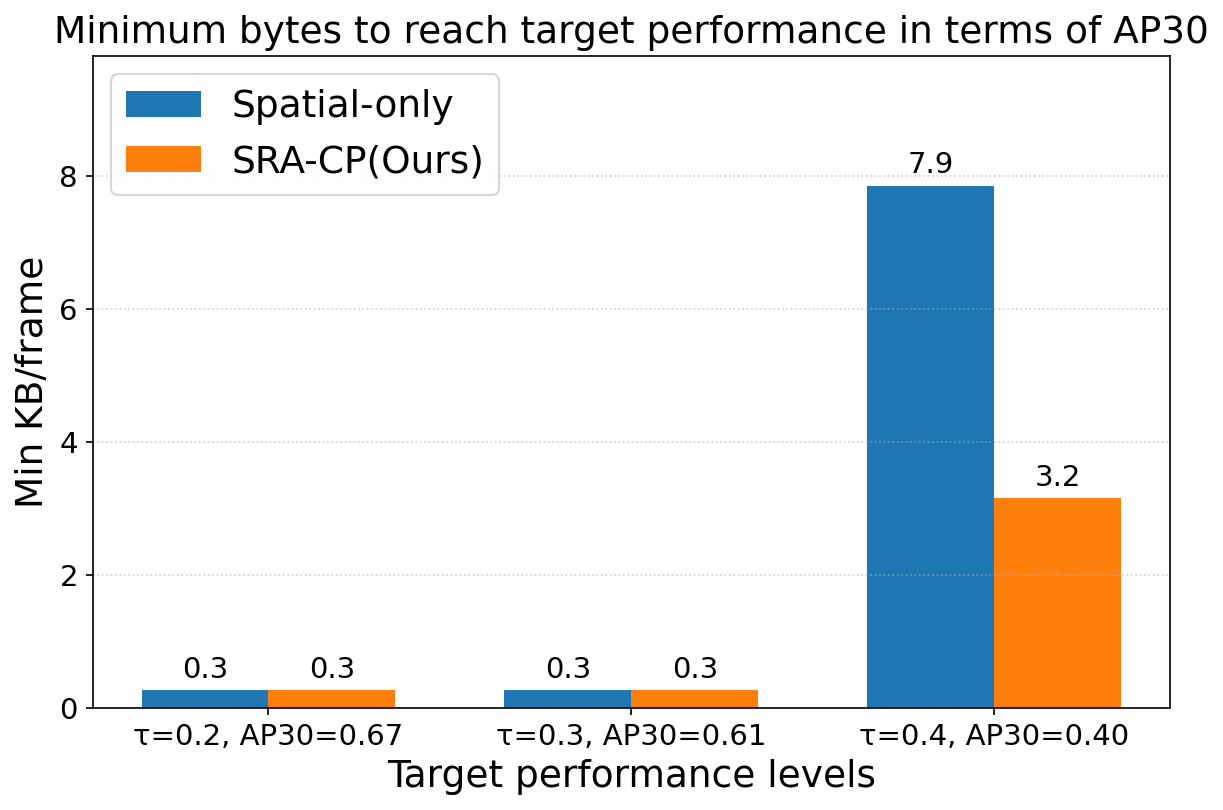}

    \end{subfigure}
    \begin{subfigure}[b]{0.32\textwidth}
        \includegraphics[width=\textwidth]{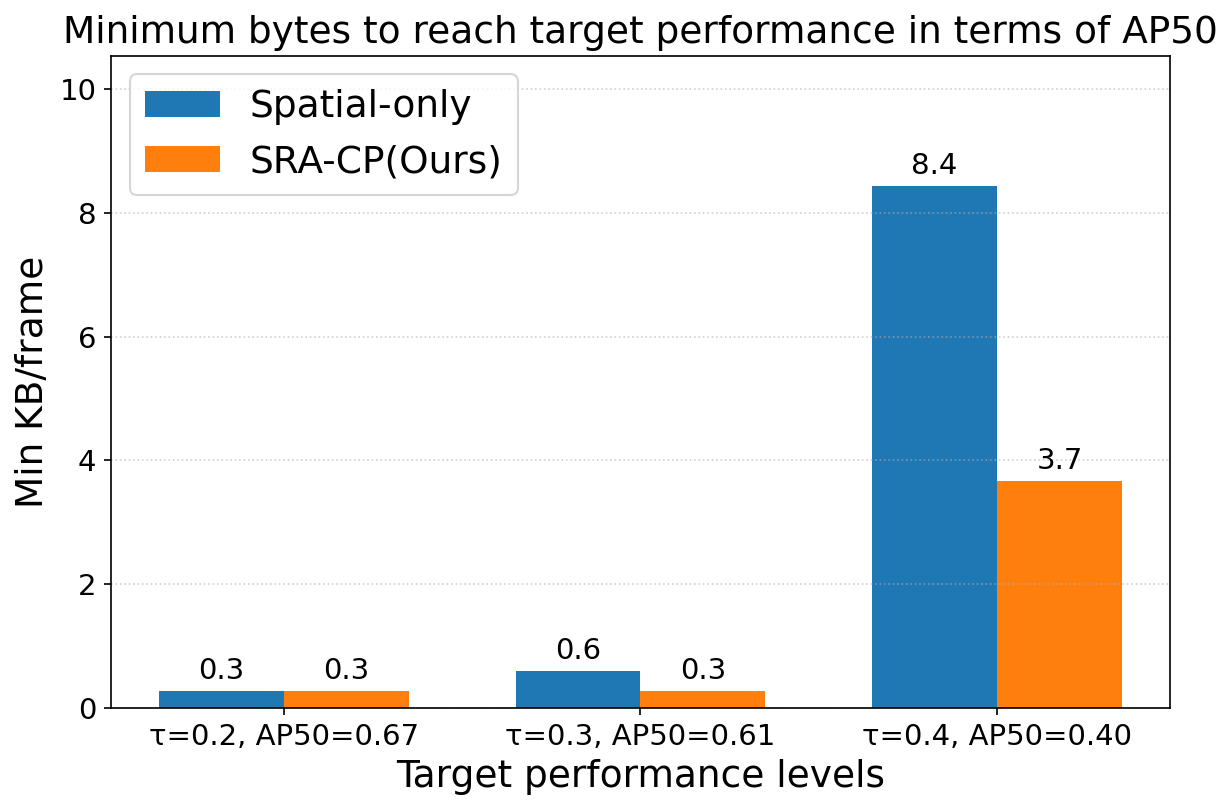}

    \end{subfigure}
    \begin{subfigure}[b]{0.32\textwidth}
        \includegraphics[width=\textwidth]{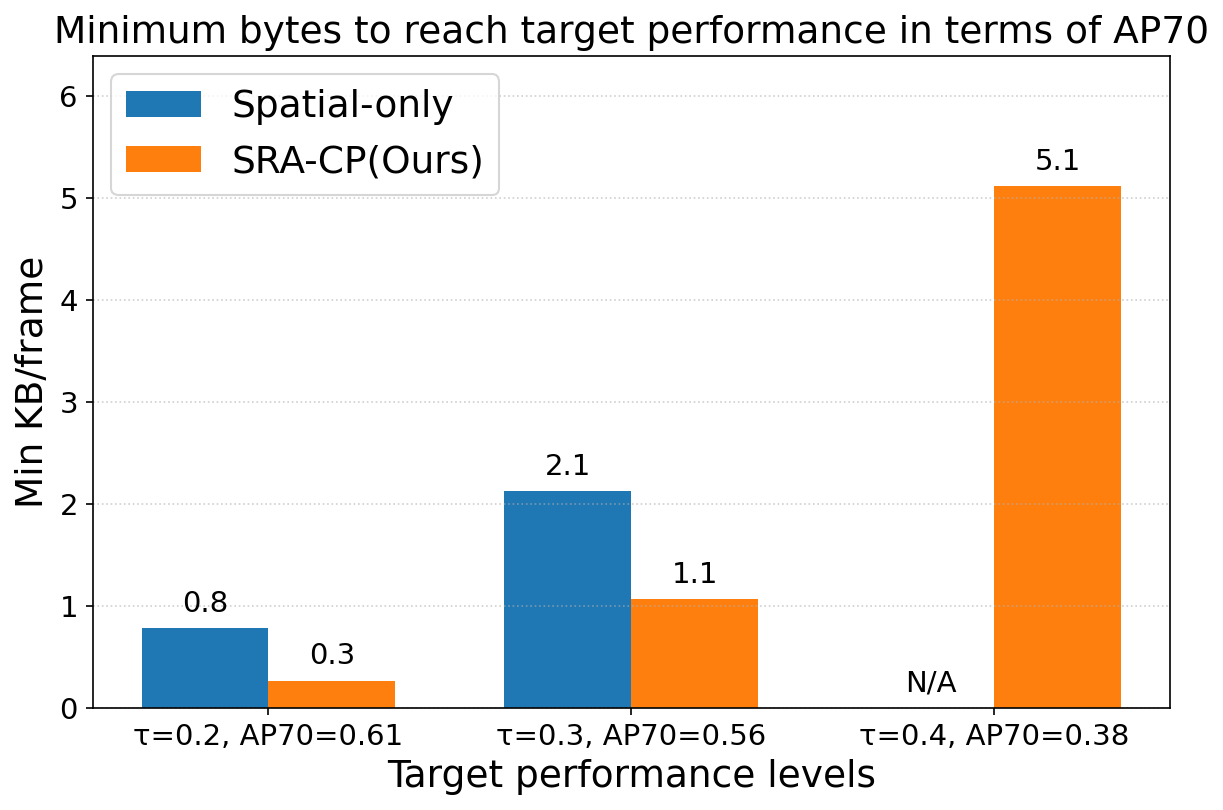}

    \end{subfigure}
    \\[1ex]

    \begin{subfigure}[b]{0.32\textwidth}
        \includegraphics[width=\textwidth]{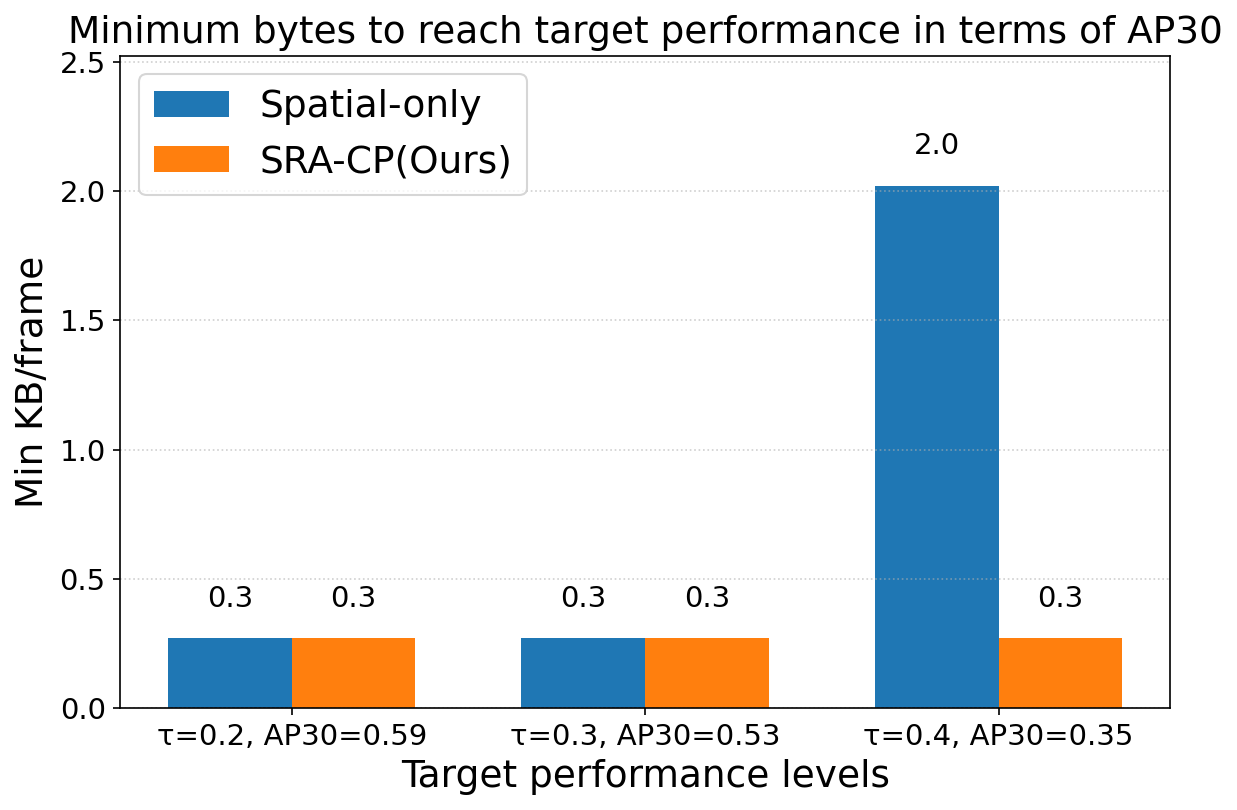}

    \end{subfigure}
    \begin{subfigure}[b]{0.32\textwidth}
        \includegraphics[width=\textwidth]{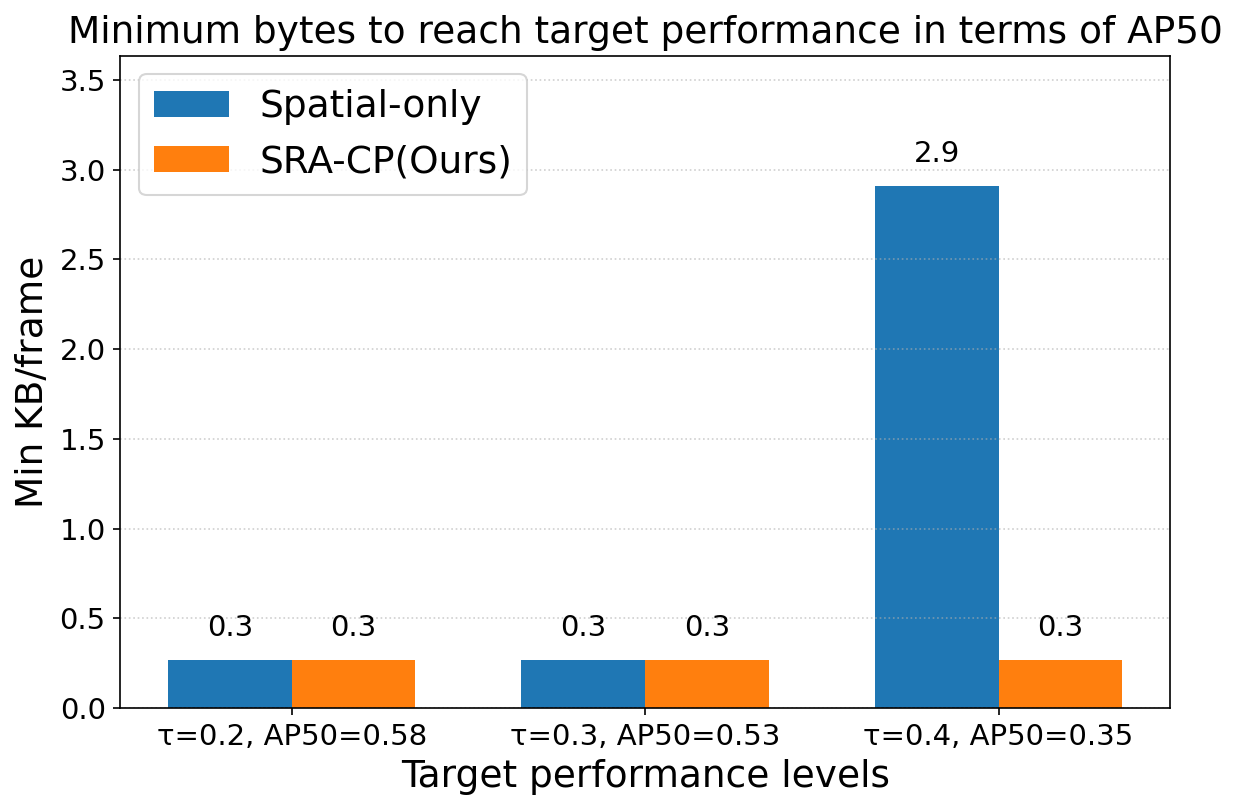}

    \end{subfigure}
    \begin{subfigure}[b]{0.32\textwidth}
        \includegraphics[width=\textwidth]{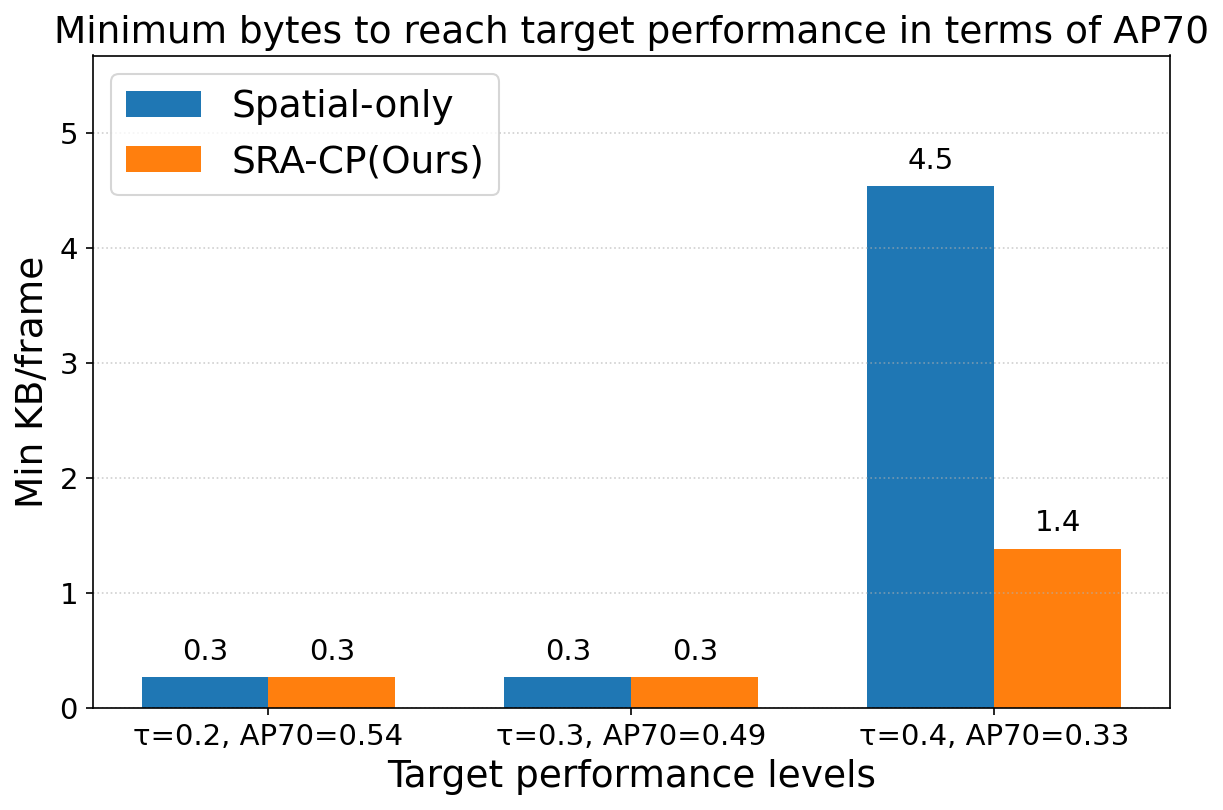}

    \end{subfigure}

    \caption{Comparison of different CP methods in terms of the minimum communication bandwidth (KB/frame) required to achieve specific perception accuracy levels (AP30, AP50, AP70) for objects with varying risk levels, categorized by risk thresholds $\tau \in \{0.2, 0.3, 0.4\}$.}
    \label{fig:min_bytes}
\end{figure}

Figure~\ref{fig:min_bytes} reports the minimum bandwidth (KB/frame) required to achieve specific Risk-AP30/AP50/AP70 targets under different risk thresholds. Across all nine subplots in Figure~\ref{fig:min_bytes}, our method consistently requires fewer bytes per frame to reach the same Risk-AP target compared to the baseline, demonstrating superior efficiency across all thresholds $\tau \in {\{0.2, 0.3, 0.4\}}$.

It is worth noting that our target values for Risk-AP were determined in a principled way: for each risk threshold $\tau$, we set the target AP values (for AP30, AP50, and AP70) to 0.9×, 0.8×, and 0.7× of the upper bound performance, respectively. This provides a reasonable and balanced target scale—stringent enough to challenge the communication strategy, yet attainable for well-designed cooperative frameworks.

For example, at $\tau = 0.2$ and AP50=0.75, our method reaches the target Risk-AP using only 1.3 KB/frame, compared to the baseline’s 3.3 KB/frame. The advantage becomes even more pronounced under higher risk thresholds: at $\tau = 0.4$ and AP70=0.42/0.38, the baseline fails to achieve the required AP target across all IoU levels (Figure~\ref{fig:min_bytes}(c, f)), while our method maintains strong performance with  5.1 and 9 KB/frame. This indicates that when perception becomes safety-critical, the baseline communication policy saturates its bandwidth without sufficient accuracy gain, whereas our SRA-CP-driven policy continues to deliver usable, risk-aware perception outputs.

\begin{table}
    \centering
    \caption{Detection performance comparison (the communication transmission volume of all the baselines are 20\% of the Upper Bound) (\textit{AP score higher is better}).}
    \begin{tabular}{lccc}
        \toprule
        \textbf{Method} & \textbf{AP30} & \textbf{AP50} & \textbf{AP70} \\
        \midrule
        Upper Bound                               & 0.9057 & 0.8955 & 0.7996 \\
        \textbf{Ours}                            & \textbf{0.8920} & 0.8731 & \textbf{0.7979} \\
        Where2comm (spatial-only)                     & 0.8902 & \textbf{0.8791} & 0.7928 \\
        Fixed-Neighbor (equal-budget, ours-union) & 0.8341 & 0.8159 & 0.6857 \\
        Random-Cell (ours-union)                  & 0.8337 & 0.8156 & 0.6861 \\
        Lower Bound                               & 0.8190 & 0.7908 & 0.6263 \\
        \bottomrule
    \end{tabular}
    \label{tab:ap_comparison}
\end{table}
\begin{table}
    \centering
    \caption{Risk-aware detection performance across risk thresholds ($\tau\!=\!0.2/0.3/0.4$). Higher is better.}
    \begin{tabular}{lcccc}
        \toprule
        \textbf{Method} & \textbf{Risk $\tau$} & \textbf{AP30} & \textbf{AP50} & \textbf{AP70} \\
        \midrule
        \multirow{3}{*}{Upper Bound}
            & 0.2 & 0.8461 & 0.8411 & 0.7745 \\
            & 0.3 & 0.7659 & 0.7632 & 0.6962 \\
            & 0.4 & 0.5003 & 0.4994 & 0.4704 \\
        \midrule
        \multirow{3}{*}{\textbf{Ours}}
            & 0.2 & \textbf{0.8365} & \textbf{0.8315} & \textbf{0.7667} \\
            & 0.3 & \textbf{0.7642} & \textbf{0.7622} & \textbf{0.6998} \\
            & 0.4 & \textbf{0.4963} & \textbf{0.4955} & \textbf{0.4702} \\
        \midrule
        \multirow{3}{*}{Where2comm (spatial-only)}
            & 0.2 & 0.8203 & 0.8136 & 0.7512 \\
            & 0.3 & 0.7412 & 0.7354 & 0.6807 \\
            & 0.4 & 0.4701 & 0.4553 & 0.4177 \\
        \midrule
        \multirow{3}{*}{Fixed-Neighbor (equal-budget)}
            & 0.2 & 0.7644 & 0.7519 & 0.6519 \\
            & 0.3 & 0.6640 & 0.6553 & 0.5705 \\
            & 0.4 & 0.3610 & 0.3565 & 0.3171 \\
        \midrule
        \multirow{3}{*}{Random-Cell}
            & 0.2 & 0.7641 & 0.7511 & 0.6505 \\
            & 0.3 & 0.6670 & 0.6578 & 0.5723 \\
            & 0.4 & 0.3737 & 0.3685 & 0.3238 \\
        \midrule
        \multirow{3}{*}{Lower Bound}
            & 0.2 & 0.7531 & 0.7357 & 0.6191 \\
            & 0.3 & 0.6483 & 0.6374 & 0.5381 \\
            & 0.4 & 0.3631 & 0.3581 & 0.3111 \\
        \bottomrule
    \end{tabular}
    \label{tab:ap_comparison_risk_all}
\end{table}

\subsection{Ablation Studies}
We ablate key communication choices like gate type (S-only, R-only, Union), blind-zone estimation (on/off) to see whether the modules of our method are actually working.

\paragraph{Gate Mode Analysis}

\begin{table}
  \centering
  \caption{Risk-aware AP at IoU=0.3/0.5/0.7 for different gate modes (5k budget) across risk thresholds $\tau$.}
  \begin{tabular}{llccc}
    \toprule
    Gate & Metric & $\tau{=}0.2$ & $\tau{=}0.3$ & $\tau{=}0.4$ \\
    \midrule
    \multirow{3}{*}{S-only}
      & AP30 & 0.7763 & 0.6714 & 0.3716 \\
      & AP50 & 0.7608 & 0.6636 & 0.3661 \\
      & AP70 & 0.6725 & 0.5963 & 0.3302 \\
    \midrule
    \multirow{3}{*}{R-only}
      & AP30 & 0.7861 & 0.6811 & 0.3959 \\
      & AP50 & 0.7731 & 0.6722 & 0.3894 \\
      & AP70 & 0.6795 & 0.6002 & 0.3544 \\
    \midrule
    \multirow{3}{*}{Union (ours)}
      & AP30 & \textbf{0.7981} & \textbf{0.7032} & \textbf{0.4128} \\
      & AP50 & \textbf{0.7866} & \textbf{0.6959} & \textbf{0.4082} \\
      & AP70 & \textbf{0.7097} & \textbf{0.6308} & \textbf{0.3742} \\
    \bottomrule
  \end{tabular}
  \label{tab:ablation_gate}
\end{table}

We compare three gate configurations under the same 5\,kB/frame bandwidth: spatial-only (S-only), risk-only (R-only), and our hybrid Union gate that integrates both spatial and risk cues as shown in Table~\ref{tab:ablation_gate}. The results in Table~\ref{tab:ablation_gate} report Risk-Aware AP at IoU=0.3/0.5/0.7 across $\tau\!\in\!\{0.2,0.3,0.4\}$.

Across all thresholds and IoU levels, the proposed Union gate consistently outperforms both S-only and R-only variants. At $\tau{=}0.3$, for instance, Union improves AP50 from 0.6636 (S-only) and 0.6722 (R-only) to 0.6959, while at $\tau{=}0.4$ the gap widens to over +4.2\% compared with S-only. Similarly, the AP70 metric rises from 0.3302 (S-only) and 0.3544 (R-only) to 0.3742. These gains demonstrate that combining spatial coverage with risk awareness yields complementary benefits—risk-only gating favors safety-critical regions but may miss peripheral context, whereas spatial-only gating ensures broader coverage but wastes bandwidth on low-risk areas.

By unifying both criteria, the Union gate adaptively allocates transmission priority based on spatial relevance and estimated collision risk, effectively balancing perception completeness and communication efficiency. This hybrid gating thus provides a more stable and risk-sensitive communication policy, enabling the system to maintain higher detection performance even as $\tau$ increases.

\paragraph{Blind-Zone Estimation}

\begin{table}
  \centering
  \caption{Risk-aware AP at IoU=0.3/0.5/0.7 for Union gate with/without blind-zone weighting (5k budget) across $\tau$.}
  \begin{tabular}{llccc}
    \toprule
    Setting & Metric & $\tau{=}0.2$ & $\tau{=}0.3$ & $\tau{=}0.4$ \\
    \midrule
    \multirow{3}{*}{Union (no blind)}
      & AP30 & 0.7912 & 0.6877 & 0.3973 \\
      & AP50 & 0.7803 & 0.6778 & 0.3912 \\
      & AP70 & 0.6948 & 0.6104 & 0.3542 \\
    \midrule
    \multirow{3}{*}{Union (blind on, ours)}
      & AP30 & \textbf{0.7981} & \textbf{0.7032} & \textbf{0.4128} \\
      & AP50 & \textbf{0.7866} & \textbf{0.6959} & \textbf{0.4082} \\
      & AP70 & \textbf{0.7097} & \textbf{0.6308} & \textbf{0.3742} \\
    \bottomrule
  \end{tabular}
  \label{tab:ablation_blind}
\end{table}

To examine whether the model benefits from explicitly prioritizing safety-critical blind areas, we conduct an ablation study on the Union gating scheme with and without blind-zone weighting under a fixed 5\,kB/frame communication budget. The results in Table~\ref{tab:ablation_blind} report Risk-Aware AP at IoU=0.3/0.5/0.7 across risk thresholds $\tau\in\{0.2,0.3,0.4\}$.

Across all IoU and risk thresholds, enabling blind-zone weighting consistently improves detection performance. Compared to the vanilla Union gate, our method achieves an average gain of $+0.7\%$, $+1.2\%$, and $+1.6\%$ for AP30, AP50 and AP70, respectively. The improvement becomes more pronounced as the risk threshold increases. For instance, at $\tau{=}0.4$, the Risk-AP70 rises from 0.3542 to 0.3742, representing a relative gain of $+5.6\%$. This pattern suggests that the proposed weighting mechanism effectively allocates communication bandwidth toward regions with higher occlusion and potential collision risk.

Qualitatively, this mechanism acts as a “safety amplifier”: when cooperative views overlap poorly or when agents observe asymmetric blind spots, the weighting function adaptively increases the transmission priority of uncertain spatial zones. As a result, even under the same bandwidth constraint, more informative features are propagated to neighboring vehicles, enhancing risk-aware perception robustness in safety-critical scenarios.

\begin{figure}
  \centering
  \begin{subfigure}{0.9\linewidth}
    \centering
    \includegraphics[width=\linewidth]{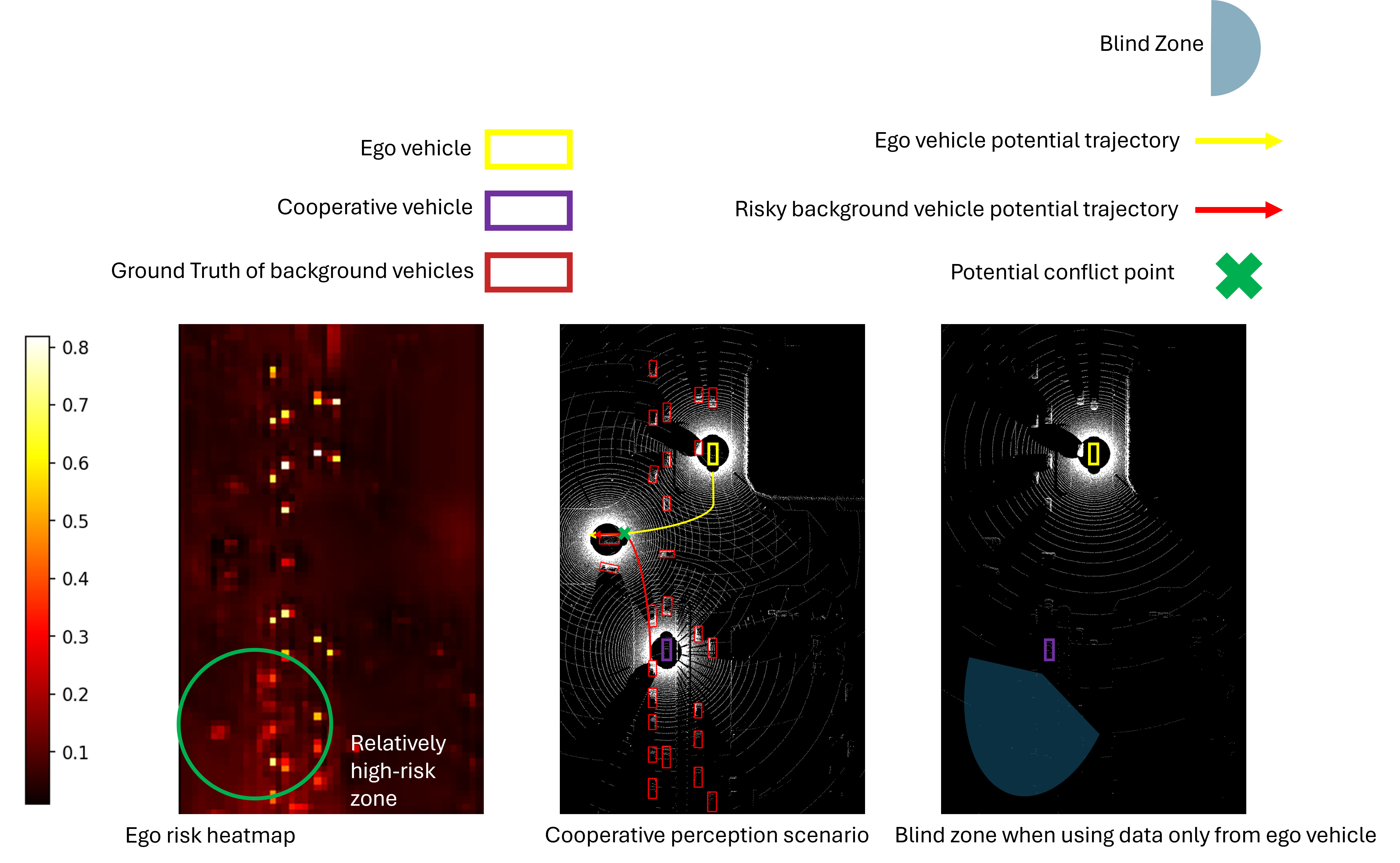}
    \caption{Predictions vs. ground truth in BEV with risk overlay.}
    \label{fig:visual_all}
  \end{subfigure}
  \vspace{0.3em}
  \begin{subfigure}{0.9\linewidth}
    \centering
    \includegraphics[width=\linewidth]{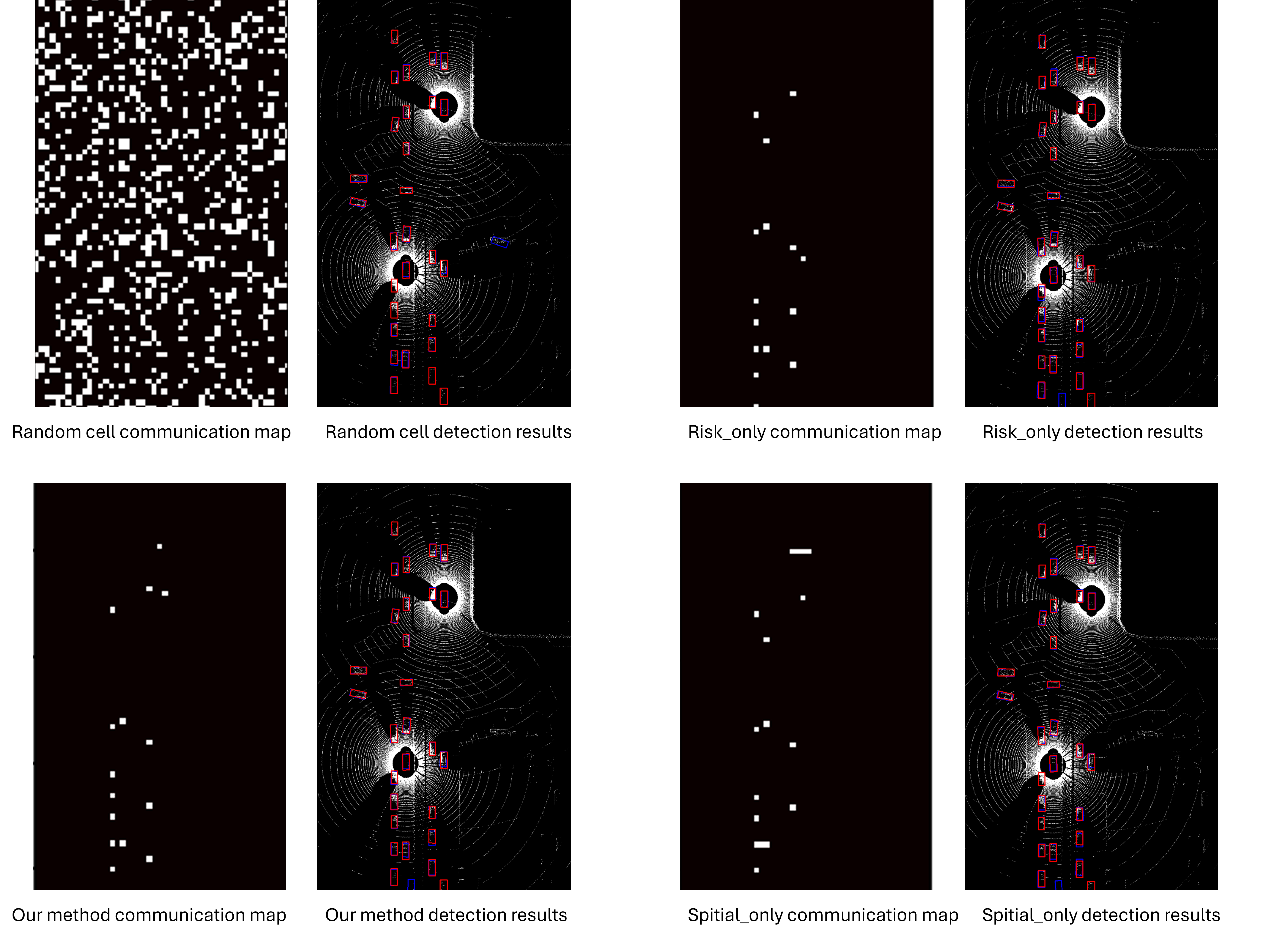}
    \caption{Risk heatmap and per‑cell transmission (Union vs. baselines).}
    \label{fig:visual_separate}
  \end{subfigure}
  \caption{Qualitative example at an unprotected intersection. Our method prioritizes risky blind‑zone cells, recovering occluded targets with fewer transmission bytes.}
  \label{fig:visual}
\end{figure}

\subsection{Visualization \& Case Study}\label{visual}
Figure~\ref{fig:visual} illustrates a challenging unprotected left-turn scenario with dense cross-traffic. The ego vehicle intends to turn left, yet its LiDAR alone cannot observe the incoming traffic hidden behind other vehicles' occlusions. These blind-zone regions coincide with locations where high-risk background vehicles are approaching, making the timely restoration of occluded agents crucial for safe maneuver planning.

We compare four communication strategies: a random-cell baseline, spatial-only, risk-only, and our Union (SRA-CP) method. The spatial-only, risk-only, and Union methods all operate under the same fixed communication budget, whereas the random baseline uses a significantly higher budget, illustrating how communication volume alone does not guarantee performance.

\paragraph{Ours vs. Spatial-only and Risk-only.}

Despite using the same byte budget, the three strategies prioritize cells differently:

Spatial-only focuses solely on geometric visibility difficulty.
It successfully identifies cells that are hard to perceive but often fails to emphasize high-risk agents located in traffic-conflict regions. As a result, it may transmit cells that are geometrically interesting yet irrelevant for imminent collision risk, while missing the truly dangerous ones.

Risk-only allocates nearly all bandwidth to the high-risk region.
This improves awareness of hazardous agents but ignores spatial fusion quality, often leading to incomplete or noisy reconstructions because difficult-to-fuse regions receive insufficient coverage.

Union (Ours) balances both spatial fusion difficulty and collision risk. It means SRA-CP suppresses low-value regions and forms a dense transmission corridor aligned with the ego–background conflict path, precisely where the occluded vehicle lies. As shown in the detection overlays, Union restores the hidden vehicle more reliably and aligns closer with the ground truth than either single-objective method.

\paragraph{Ours vs. Random-cell Communication.}
Even with a much larger number of transmitted cells, the random-cell method performs poorly. Because cells are sampled uniformly at random, it often allocates bandwidth to irrelevant free-space areas while failing to cover the critical blind-zone region at the correct moment. Consequently, the recovered detection remains incomplete or inconsistent despite the inflated budget.

In contrast, Union (SRA-CP) pinpoints and transmits only the essential cells—those that influence collision risk or improve multi-agent fusion quality—and thus reconstructs the critical occluded vehicle with dramatically fewer bytes.

\paragraph{What Gets Transmitted (Per-cell Transmission Maps).}

The transmission maps further confirm each method’s behavior:

Spatial-only spreads bytes broadly across many cells—high coverage but low efficiency.
Risk-only over-concentrates in a compact region—high focus but weak contextual support.
Random shows noisy, unstructured coverage even with a large budget—no semantic prioritization.
Union (Ours) exhibits an intelligent, elongated high-density band that tracks the potential collision trajectory while maintaining minimal peripheral context.

This pattern matches the ablation findings in Table~\ref{tab:ablation_gate}: combining spatial difficulty and risk factors yields the most efficient allocation strategy.

In summary, our experiments show that SRA-CP consistently dominates existing cooperative-perception baselines in the communication–safety trade-off. Under the same communication budget, SRA-CP matches or exceeds the cutting-edge spatial-only selective method in perception accuracy, while delivering notably higher perception accuracy for safety-critical objects. When sweeping the per-link budget, our method traces the Pareto frontier: for any given bandwidth it attains the best risky-object detection, and for any target perception accuracy it requires substantially fewer transmitted bytes than competing schemes. Qualitative case studies at unprotected intersections further illustrate that SRA-CP automatically concentrates messages on risky blind-zone cells, allowing the ego vehicle to recover occluded, dangerous agents earlier and more reliably during driving.

\section{Conclusion}

This paper presents a novel Spontaneous Risk-Aware Selective Cooperative Perception (SRA-CP) framework to address the scalability and bandwidth challenges of multi-agent cooperative perception in dynamic traffic environments. We first design a protocol in which connected agents continuously broadcast their perception coverage with very low communication cost and initiate on-demand handshakes when risk-relevant blind zones are detected. For a certain connected agent, we propose a perceptual risk identification model to detect and quantify risk-critical occlusions, a selective information sharing model to determine which features to transmit under bandwidth constraints, and a dual-attention feature fusion model to integrate received features into the ego agent’s perception output. 

Extensive evaluations on a public dataset were conducted against five baseline methods, each targeting a different aspect of the problem. These include a cutting-edge selective CP method, a fully connected CP setting as an upper bound, a no-CP setup as a lower bound, and another 2 methods: fixed neighbor allocation and random feature Sampling to evaluate the effects of communication target selection and content-level feature prioritization, respectively. Experimental results show that SRA-CP achieves less than 1\% loss for safety-critical objects compared to generic CP, while using only 20\% of the communication bandwidth.  Moreover, compared to the cutting-edge selective CP method, SRA-CP improves the AP for critical objects by 15\% under the same bandwidth budget, demonstrating its communication efficiency and risk-awareness advantage.

As future work, we are collecting real-world driving data using our lab's connected vehicles. We plan to further evaluate the framework on this in-house dataset and conduct field tests to assess its real-world applicability and robustness.

\bibliographystyle{plainnat} 
\bibliography{ref}

\end{document}